\documentclass[journal]{IEEEtran}

\usepackage{cite}
\usepackage{amsmath,amssymb,amsfonts}
\usepackage{algorithmic}
\usepackage{graphicx}
\usepackage{textcomp}
\usepackage{xcolor}
\usepackage{times}
\usepackage{soul}
\usepackage[utf8]{inputenc}
\usepackage{amsmath}
\usepackage{amsfonts}
\usepackage{multirow}
\usepackage{amsthm}

\usepackage{algorithm} 
\usepackage{caption,subcaption}
\usepackage[justification=centering]{subcaption}
\usepackage{color}
\usepackage{mathtools}
\usepackage{mathptmx}
\usepackage{float}
\usepackage{color}
\usepackage{makecell}
\usepackage{epstopdf}
\usepackage{colortbl}
\usepackage{pdflscape}
\usepackage{booktabs}
\usepackage{threeparttable}
\usepackage{makecell}
\usepackage{hyperref}
\usepackage{tablefootnote}
\usepackage{footnotehyper}
\def\BibTeX{{\rm B\kern-.05em{\sc i\kern-.025em b}\kern-.08em
		T\kern-.1667em\lower.7ex\hbox{E}\kern-.125emX}} 
\begin{document}
\title{Pro-UIGAN: Progressive Face Hallucination from Occluded Thumbnails}

\author{Yang Zhang, Xin Yu, Xiaobo Lu, Ping Liu
\thanks{Y. Zhang is with the Key Laboratory of Intelligent Perception and Systems for High-Dimensional Information, Ministry of Education, School of Computer Science and Engineering, Nanjing University of Science and Technology, Nanjing 210094, China, e-mail: zhangyang201703@126.com.

X. Yu is with the Australian Institute of Artificial Intelligence, University of Technology Sydney, Ultimo, NSW 2007, Australia, e-mail: xin.yu@uts.edu.au.

X. B. Lu is with the School of Automation, Southeast University, Nanjing 210096, China; Key Laboratory of Measurement and Control of Complex Systems of Engineering, Ministry of Education, Nanjing 210096, China, e-mail: xblu2013@126.com.

P. Liu (corresponding author) is with the Center for Frontier AI Research (CFAR), Research Agency for Science, Technology and Research (A*STAR), Singapore 138634, e-mail: pino.pingliu@gmail.com.
}
}

\markboth{Journal of \LaTeX\ Class Files,~Vol.~14, No.~8, August~2015}%
{Shell \MakeLowercase{\textit{et al.}}: Bare Demo of IEEEtran.cls for IEEE Journals}
\maketitle

\begin{abstract}
In this paper, we study the task of hallucinating an authentic high-resolution (HR) face from an occluded thumbnail.
We propose a multi-stage Progressive Upsampling and Inpainting Generative Adversarial Network, dubbed Pro-UIGAN, which exploits facial geometry priors to replenish and upsample ($8\times$) the occluded and tiny faces ($16\times 16$ pixels).
Pro-UIGAN iteratively (1) estimates facial geometry priors for low-resolution (LR) faces and (2) acquires non-occluded HR face images under the guidance of the estimated priors.
Our multi-stage hallucination network upsamples and inpaints occluded LR faces via a coarse-to-fine fashion, significantly reducing undesirable artifacts and blurriness.
Specifically, we design a novel cross-modal attention module for facial priors estimation, in which an input face and its landmark features are formulated as queries and keys, respectively.
Such a design encourages joint feature learning across the input facial and landmark features, and deep feature correspondences will be discovered by attention.
Thus, facial appearance features and facial geometry priors are learned in a mutually beneficial manner.
Extensive experiments show that our Pro-UIGAN attains visually pleasing completed HR faces, thus facilitating downstream tasks, \emph{i.e.,} face alignment, face parsing, face recognition as well as expression classification.
\end{abstract}


\IEEEpeerreviewmaketitle

\section{Introduction}
With the increasing demand for social security, non-intrusive identity verification has become indispensable in daily life. 
As the face is one of the most frequently utilized biometric cues, it is highly desirable to obtain face images of high quality for providing essential information for identity verification.
However, in real scenarios, the captured faces might not only be in low resolutions due to the long distance but also undergo occlusions caused by body parts or accessories, such as eyeglasses, scarves, etc.
Due to the low resolution and occlusions, it becomes difficult, if not impossible, to extract useful knowledge to support downstream applications, such as face verification and facial attribute classification.
Therefore, it becomes necessary to design advanced methods to hallucinate LR images with various occlusions.

\begin{figure}[t]
\centering
\includegraphics[height=10cm,width=0.48\textwidth]{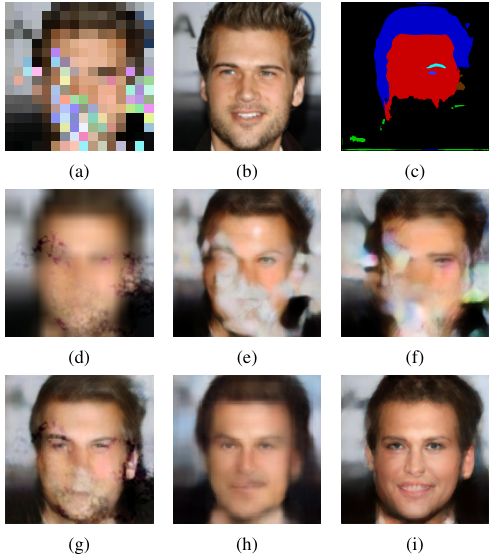}
\caption{Face inpainting and upsampling results on an occluded LR face. (a) The $16\times 16$ occluded LR face image. (b) The original $128\times 128$ non-occluded HR face image. (c) Face parsing result of (a), which is obtained by applying bicubic interpolation followed by the work~\cite{Liu_2015_CVPR}. 
(d) Face inpainting result of (a), which is obtained by applying bicubic interpolation followed by the work~\cite{li2017generative}.
(e) Face SR result of (a) by applying~\cite{yu2018face}. (f) Result obtained by applying~\cite{li2017generative} followed by~\cite{yu2018face}. (g) Result obtained by applying~\cite{yu2018face} followed by~\cite{li2017generative}. (h) Result of FCSR-GAN~\cite{cai2019fcsrj}. (i) Result of our Pro-UIGAN.}
\label{fig1}
\end{figure}

One straightforward solution to generate corresponding HR face images without occlusions from the LR face images with occlusions is to directly combine existing face SR methods~\cite{zhu2016deep,cao2017attention,chen2018fsrnet,yu2018imagining,yu2018face,menon2020pulse} and face inpainting solutions~\cite{li2017generative,2018Geometry}, \textit{i.e.}, inpaint + SR, or SR + inpaint.
However, this straightforward sequentially combined solution might be sub-optimal due to the following reasons: On the one hand, off-the-shelf face inpainting methods usually require a high-resolution (HR) face image as their inputs to analyze the context to ``fill in" the missing region.
When the input images are in low resolutions, for example, $16 \times 16$ pixels, accurately identifying different facial regions (see Fig.~\ref{fig1}(c)) for the context extraction becomes impractical.
In such a case, directly inpainting an LR image with occlusions leads to blurred facial details (see Fig.~\ref{fig1}(d)), which might be exaggerated in the following face SR process (see Fig.~\ref{fig1}(f)).
On the other hand, most existing face SR works only consider non-occluded face images as their inputs.
These methods may suffer from ghosting artifacts in the results when LR inputs with occlusions are given (see Fig.~\ref{fig1}(e)).
Sequentially, when we apply face inpainting methods on those images with artifacts, the final results will exhibit severe distortions (see Fig.~\ref{fig1}(g)).
Based on the analysis and experimental result observations, we believe that: treating face SR and face inpainting as two independent tasks and handling them sequentially and independently may not be an ideal solution.

Very recently,~\cite{cai2019fcsrj} proposes a deep generative adversarial network (FCSR-GAN) to jointly super-resolve and inpaint occluded face images in low resolutions.
Sharing a similar spirit with us, FCSR-GAN achieves better performance by considering the relation between two tasks, \textit{i.e.}, face inpainting, face super-resolution.
However, without a progressive processing mechanism, FCSR-GAN inpaints occluded LR faces in only one single stage and then super-resolves the non-occluded LR faces by a large magnification factor ($8\times$ or $16\times$) in another solo-stage.
Apparently, as a ``one-shot deal” method, FCSR-GAN
does not possess a ``looking back” mechanism to refine the results iteratively.
As demonstrated in Fig.~\ref{fig1}(h), although FCSR-GAN achieves better results than prior works, over-smoothed facial details and severe distortions still exist in the results generated by FCSR-GAN~\cite{cai2019fcsrj}.

In this paper, we propose to hallucinate occluded LR inputs$\footnote[1]{occluded LR faces: low-resolution faces with occlusions.}$ while achieving non-occluded HR faces$\footnote[2]{non-occluded HR faces: high-resolution faces without occlusions.}$ in a unified framework.
In this manner, these two tasks (\emph{i.e.,} face SR and face inpainting) are addressed simultaneously and facilitate each other mutually.
Compared to FCSR-GAN, our joint mechanism is accomplished by a multi-stage progressive hallucination strategy.
We design a progressive joint face SR and face inpainting framework, dubbed Pro-UIGAN.
Pro-UIGAN runs in a multi-stage manner where each stage refines the face images hallucinated at the previous stage.
By iteratively refining the details of hallucinated images, we can achieve high-quality results for large upsampling factors.
Not only that, equipped with a delicate designed cross-modal attention mechanism, we exploit facial geometry priors (\emph{i.e.}, facial landmark heatmaps) as the semantic guidance during our progressive hallucination process to reconstruct more realistic facial details.

Our Pro-UIGAN consists of a Pro-UInet and two discriminators, \emph{i.e.,} a local discriminator (namely Local-D) and a global discriminator (namely Global-D).
The Pro-UInet stacks a few successive Upsampling and Inpainting Blocks (UI-blocks) and generates a non-occluded HR face image by performing multiple inpainting and upsampling of an occluded LR face.
By doing so, we can inpaint and super-resolve the LR input in a coarse-to-fine manner.
In particular, in each stage, our UI-block comprises a Cross-modal Attention Module (CM-AM) and a Transformative Upsampling Module (TUM).
It estimates the most distinguishable facial landmarks in an input face and constructs facial geometry priors to guide face hallucination.
Our designed CM-AM employs the input face and its landmark features as queries and keys to calculate their cross-attention matrix for deep feature correspondences construction.
Then, it generates facial geometry priors and facial appearance features in a mutual promotion manner.
In our multi-stage hallucination process, the first UI-block generates a coarse hallucinated face from the occluded LR input, providing guidance for following UI-blocks.
Then, the following UI-blocks refine the face images hallucinated at the previous stage and generate hallucination results with finer details.
Fig.~\ref{fig1}(i) illustrates that our hallucinated non-occluded HR face is more photo-realistic than the results of the state-of-the-art.

Overall, our contributions are threefold:
\begin{itemize}
\item We present a novel framework, namely Progressive Upsampling and Inpainting Generative Adversarial Network (Pro-UIGAN), to jointly achieve face inpainting and face SR in a unified framework. We design a multi-stage hallucination and inpainting interwoven strategy. Specifically, we upsample and inpaint occluded LR faces in a coarse-to-fine fashion, thereby reducing undesirable artifacts and noises compared to a direct combination of face hallucination and inpainting methods.
\item We propose a Cross-Modal Attention Module (CM-AM) to learn facial geometry priors and facial appearance features collaboratively.
Our CM-AM provides effective clues for feature alignment and enhancement and thus promotes more accurate face hallucination.
\item Extensive experiments manifest that our Pro-UIGAN authentically replenishes ($8\times$ upsampling) occluded and LR face images (\emph{i.e.,} $16\times 16$ pixels).
Moreover, our Pro-UIGAN provides superior hallucinated face images for downstream tasks, \emph{i.e.,} face alignment, face parsing, face recognition as well as expression classification.
\end{itemize}

\section{Related Work}
\subsection{Face Super-resolution}
Face Super-resolution (SR) aims at enhancing the resolution of LR face images to generate corresponding HR face images.
The previous works can be generally grouped into three categories: holistic-based~\cite{wang2005hallucinating,liu2007face,kolouri2015transport}, part-based~\cite{ma2010hallucinating,jiang2014face,li2014face,farrugia2017face,liu2017robust,tappen2012bayesian,yang2018hallucinating}, as well as deep learning-based solutions~\cite{yu2016ultra,yu2018face,huang2019wavelet,yu2017face,cao2017attention,xu2017learning,dahl2017pixel,oord2016pixel,yu2019hallucinating,zhang2020copy}.

Holistic-based methods employ global face models to upsample LR faces.
Wang~\textit{et al.}~\cite{wang2005hallucinating} establish a linear mapping between LR and HR images to achieve face SR based on an Eigen-transformation of LR faces.
Liu~\textit{et al.}~\cite{liu2007face} incorporate bilateral filtering to mitigate the ghosting artifacts, improving the quality of generated HR faces.
Kolouri and Rohde~\cite{kolouri2015transport} introduce optimal transport and subspace learning to morph HR faces from aligned LR ones.
However, they all require LR face images to be aligned precisely, and reference HR faces are under canonical poses and natural expressions.

To handle large poses and complex expressions, part-based methods have been proposed to super-resolve local facial regions rather than enforcing global constraints.
The works~\cite{ma2010hallucinating,li2014face,jiang2014face,farrugia2017face} incorporate facial patches extracted from HR datasets to enhance input LR facial regions.
Marshall~\textit{et al.}~\cite{tappen2012bayesian} employ the SIFT Flow algorithm to warp exemplar faces and compute the hallucinated HR image through MAP estimation.
Yang~\textit{et al.}~\cite{yang2018hallucinating} use facial landmarks to retrieve adequate HR facial component exemplars for further face SR.
Those works~\cite{tappen2012bayesian,yang2018hallucinating} need to precisely localize facial components for face SR, which is challenging, especially in LR cases.

Benefiting from the strong feature extraction ability of deep neural networks, deep learning-based solutions achieve promising performance compared to prior methods~\cite{wang2005hallucinating,liu2007face,kolouri2015transport,ma2010hallucinating,jiang2014face,li2014face,farrugia2017face,liu2017robust,tappen2012bayesian,yang2018hallucinating}.
Yu \textit{et al.}~\cite{yu2016ultra} design a GAN-based model to upsample LR faces.
Huang \textit{et al.}~\cite{huang2019wavelet} introduce wavelet coefficients into CNNs to super-resolve LR faces with multiple upscaling factors.
Cao \textit{et al.}~\cite{cao2017attention} put forward an attention-aware mechanism and a local enhancement network to progressively enhance local facial regions during hallucination.
Xu \textit{et al.}~\cite{xu2017learning} exploit a multi-class adversarial loss to promote joint face SR and deblurring.
Dahl \textit{et al.}~\cite{dahl2017pixel} propose to super-resolve pre-aligned LR inputs via designing an autoregressive Pixel-RNN~\cite{oord2016pixel}.
Yu \textit{et al.}~\cite{yu2018face} incorporate facial component information from the intermediate upsampled features into an upsampling stream to achieve superior face hallucination results.
Yu \textit{et al.}~\cite{yu2019hallucinating} present a multiscale transformative discriminative neural network to super-resolve unaligned and very small faces of variable resolutions.
Zhang \textit{et al.}~\cite{zhang2020copy} present a two-branch upsampling framework to normalize and super-resolve non-uniform illumination and LR inputs.
Menon \textit{et al.}~\cite{menon2020pulse} present a Photo Upsampling via Latent Space Exploration (PULSE) algorithm to generate high-quality frontal face images at large resolutions.
Since those works~\cite{yu2016ultra,yu2017face,yu2018face,huang2019wavelet,cao2017attention,xu2017learning,dahl2017pixel,menon2020pulse,oord2016pixel,yu2019hallucinating,zhang2020copy} aim to super-resolve LR faces without occlusions, they might obtain inferior results when occluded LR inputs are given, as seen in Fig.~\ref{fig1}(e).

\subsection{Image Inpainting}
Image inpainting is to generate missing regions in corrupted images.
The inpainted images should be not only visually realistic but also consistent in content.
Image inpainting techniques can be grouped into three classes.
The first class employs the diffusion equation to iteratively propagate low-level features from the content area to the missing region along the boundaries~\cite{bertalmio2000image,elad2005simultaneous}.
The methods belonging to the second class are patch-based methods, which search similar patches from exemplar image databases or the original image to fill in the missing contents~\cite{darabi2012image,bertalmio2003simultaneous,criminisi2004region,hays2007scene}.
The third class is learning-based methods, which employ encoder-decoder networks to extract image features and fill missing content based on the extracted features~\cite{pathak2016context,iizuka2017globally,yu2018generative,banerjee2020hallucinating}.
However, these methods generally focus on inpainting natural images rather than class-specific images, such as faces.

Face inpainting is even more challenging because the inpainting process must retain the facial topological structure and the face identity, and humans are very sensitive to distorted facial structure.
In general, researchers usually introduce facial prior information to inpainting.
S. Zhang \textit{et al.}~\cite{zhang2017demeshnet} aim to recover face images from structural obstructions such as streaks.
However, their method is more effective when only a small area is missing.
Li \textit{et al.}~\cite{li2017generative} propose a face inpainting GAN which introduces facial geometry as semantic regularization to guarantee face topological structure.
Meanwhile, some works employ global and local discriminators to ensure the reality of completed results.
However, their method cannot handle expression and pose variations and may fail when meeting LR occluded faces.
\cite{2018Geometry} designs a geometry-aware face inpainting model which exploits facial geometry information as guidance for inpainting.
Liu \textit{et al.}~\cite{liu2017semantically} integrate perceptual subnetwork to capture semantic-level facial features, thus improving synthesized content details.
While the solutions~\cite{2018Geometry,liu2017semantically} concentrate on guaranteeing the integrity of image structure, they neglect the reality of the texture of critical image areas (\emph{e.g.,} key facial components).
Zhou \textit{et al.}~\cite{zhou2020learning} propose to learn the relationships between multiple-scale facial textures and generate facial priors based on the location of facial components for face inpainting.
Although~\cite{zhou2020learning} can produce high-fidelity face images with fine-grained facial components, it still suffers from severe performance degradation when inputs are in low resolutions.

\section{Methodology}
\begin{figure*}[t]
\centering
\includegraphics[height=12cm,width=0.95\textwidth]{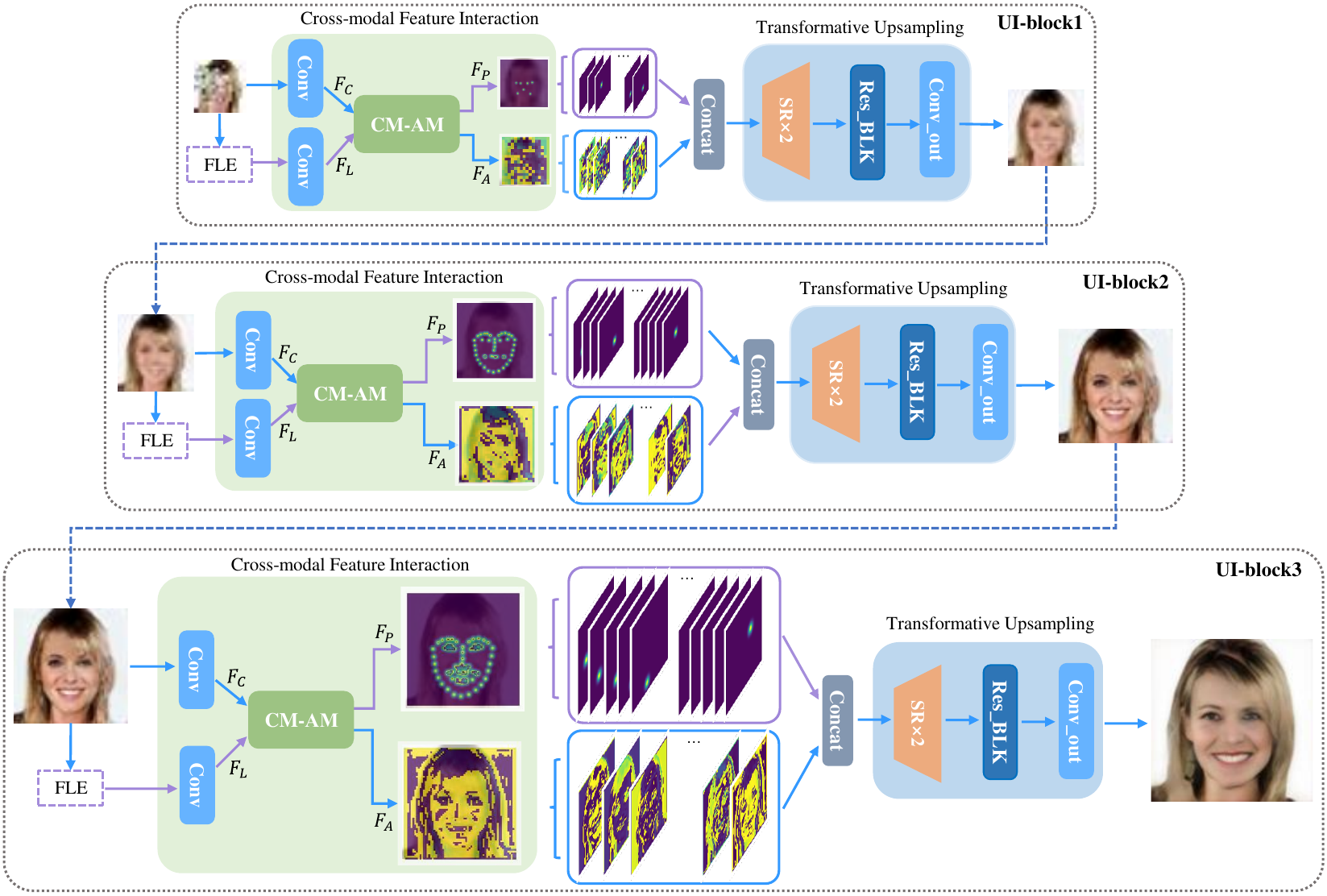}
\caption{The architecture of our Pro-UInet. Here, FLE represents a facial landmark estimation model~\cite{bulat2017far}. CM-AM represents the cross-modal attention module.}
\label{fig3}
\end{figure*}

In this section, we illustrate the technical details of the proposed Progressive Upsampling and Inpainting Generative Adversarial Network (Pro-UIGAN).
Our Pro-UIGAN is designed based on a multi-stage progressive hallucination strategy.
It comprises a Pro-UInet and two discriminators, \emph{i.e.,} Local-D and Global-D.
The Pro-UInet estimates facial geometry priors, \emph{i.e.,} facial landmark heatmaps, and uses them as the semantic guidance to obtain non-occluded HR faces progressively.
Meanwhile, the Local-D and Global-D enforce the hallucinated faces to be photo-realistic.
The whole pipeline is shown in Fig.~\ref{fig3}.

\subsection{Multi-stage Progressive Hallucination Strategy}
Inspired by progressive curriculum learning~\cite{bengio2009curriculum, karras2018progressive,morerio2017curriculum,korshunova2017fast,ulyanov2016texture}, we propose a multi-stage progressive hallucination strategy, where our network might be weak initially and keep improving stage by stage.
In the first stage, our network produces a coarse hallucinated face from an occluded LR input, providing guidance for the following stage.
In the following stage, our network refines the result provided by the previous stage.
The refined result provides more delicate knowledge for the next stage.
Consequently, our network generates a photo-realistic HR face in a coarse-to-fine manner.
We demonstrate that our proposed progressive learning strategy and designed network significantly reduce blurriness and artifacts in the hallucinated face image, as shown in Fig.~\ref{figloss}(l).

\subsection{Pro-UInet}
In line with our multi-stage progressive hallucination strategy, our Pro-UInet stacks a series of Upsampling and Inpainting Blocks (UI-blocks) and reconstructs non-occluded HR face images progressively (see Fig.~\ref{fig3}).

First, we send an occluded LR face and its facial landmark features estimated by~\cite{bulat2017far} into the first UI-block.
Our UI-block comprises a Cross-Modal Attention Module (CM-AM) and a Transformative Upsampling Module (TUM).
The CM-AM employs the input face and its landmark features as queries and keys and learns facial geometry priors and facial appearance features collaboratively.
The TUM uses residual blocks~\cite{he2016deep} and deconvolutional layers to enhance high-frequency facial details and upsample facial feature maps.
Here, we concatenate the learned facial geometry priors and facial appearance features and send them into the TUM for feature alignment, aggregation, and $2\times$ upscaling.
As a result, the first UI-block generates a coarse completed and upsampled face (see Fig.~\ref{figloss}(c)).
Subsequently, the latter UI-blocks further inpaint and upsample the coarse hallucinated face, generating finer hallucinated ones (see Figs.~\ref{figloss}(d) and (l)).
Specifically, as the stage number increases, the input for the current stage, which is generated in the past stage, becomes more accurate in structures and provides more detailed prior knowledge to benefit the following learning processes.

\subsection{Cross-modal Attention Module (CM-AM)}
We design the CM-AM to mine face restoration clues from two different but complementary aspects: facial geometry priors (\emph{i.e.,} facial landmark heatmaps) and facial appearance features.
The facial geometry priors ($F_{P}$) represent the facial shape information, and the facial appearance features ($F_{A}$) contain the facial texture and color information.

\begin{figure}[t]
\centering
\includegraphics[height=4cm,width=0.48\textwidth]{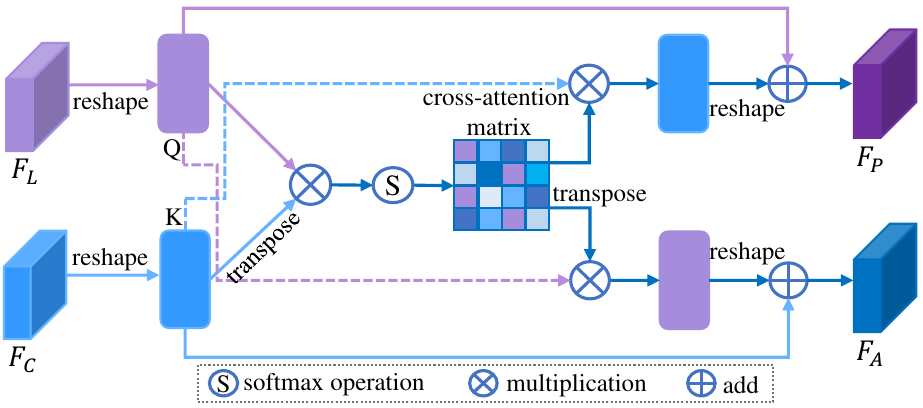}
\caption{The diagram of CM-AM. $F_{L}$, $F_{C}$, $F_{P}$, and $F_{A}$ represent facial landmark features, input facial features, facial geometry priors, and facial appearance features, respectively.}
\label{fig4}
\end{figure}

As shown in Fig.~\ref{fig4}, the facial landmark features ($F_{L}$) and the input facial features ($F_{C}$) are formulated as queries (Q) and keys (K) to calculate a cross-attention matrix (CM)~\cite{yang2020learning, qian2020thinking}.
The CM calculates the relevance between Q and K by the normalized inner product:
\begin{equation}
{CM} =\left\langle\frac{Q}{\left\|Q\right\|}, \frac{K}{\left\|K\right\|}\right\rangle.
\label{eq1}
\end{equation}
Then, the CM augments the attended features from one stream to another.
Such a design encourages joint feature learning across the input facial and landmark features, and their deep feature correspondences will be discovered by attention.
Specifically, during learning $F_{P}$ and $F_{A}$, $F_{C}$ and $F_{L}$ are formulated as the values (V), respectively.
Finally, $F_{P}$ and $F_{A}$ are learned by the following formulation:
\begin{equation}
\left\{\begin{aligned}
F_{P}=\left[{Conv}\left({Conv}(F_{C})\right)^{R} \odot CM\right]^{R} +F_{L}, \\
F_{A}=\left[{Conv}\left({Conv}(F_{L})\right)^{R} \odot CM\right]^{R} +F_{C},
\end{aligned}\right.
\label{eq2}
\end{equation}
where $Conv$ represents a convolutional layer, $\odot$ donates an element-wise multiplication operation, and $R$ is a reshape function.

To illustrate the effect of our CM-AM, we conduct experimental comparisons and report the results in Figs.~\ref{figloss}(e) and (h).
In the subfigures, we can find that the Pro-UIGAN variant without CM-AM produces inferior results.
Therefore, our CM-AM provides effective facial geometry priors as clues for feature alignment and enhancement and thus promotes more accurate face hallucination results (see Fig.~\ref{figloss}(l)).

\subsection{Transformative Upsampling Module (TUM)}
Then, we concentrate the learned facial geometry priors ($F_{P}$) and facial appearance features ($F_{A}$) and send them into the TUM for super-resolving facial details.
We use a deconvolutional layer ($H_{up}$) to upsample the concentrated feature maps ($2\times$ upscaling).
During this procedure, the facial geometry priors provide spatial configuration of facial components and shapes, guiding joint face inpainting and super-resolution.
Then, inspired by the high frequency residual learning~\cite{cheng2020high}, we adopt residual blocks~\cite{he2016deep} ($H_{res}$) to enhance the high-frequency facial details.
The processing procedure can be represented as:
\begin{equation}
F_{out} = {H_{res}}\left[{H_{up}}({Concat}(F_{P}, F_{A}))\right].
\label{eq4a}
\end{equation}

As a result, TUM not only super-resolves the facial features but also inpaints the occluded areas.
Finally, we transform the result of TUM ($F_{out}$) to the original image space and generate the hallucinated face.

\subsection{Local-Global Discriminators}
We employ two discriminators, \emph{i.e.,} Global-D and Local-D, to force the generated HR faces to lie on the same manifold as real HR faces do.
Considering that our masks may be irregular, we apply the Local-D to the central facial area, containing the nose, mouth, and eyes.
In this manner, the Local-D encourages the generated facial areas to be semantically valid, while the Global-D encourages the generated local area consistent with the other areas semantically and spatially.

\begin{figure}[t]
\centering
\includegraphics[height=8.5cm,width=0.48\textwidth]{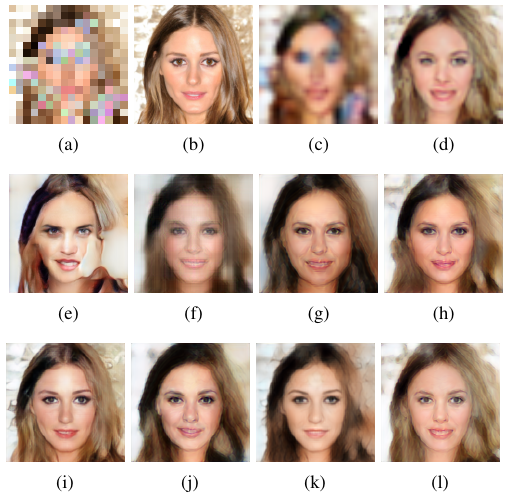}
\caption{Impacts of different components and losses on model performance. (a) Occluded LR images ($16\times 16$ pixels). (b) Ground-truth HR images ($128\times 128$ pixels). (c) Results of UI-block1. (d) Results of UI-block2. (e) Results of Pro-UIGAN without adopting CM-AMs. Note that the facial geometry is ambiguous. (f) Results of Pro-UIGAN trained by $L_{mse}$ and $L_{h}$. (g) Results of Pro-UIGAN trained by $L_{mse}$, $L_{f}$ and $L_{h}$. (h) Results of Pro-UIGAN without adopting the CM-AM in UI-block1. Here, we only employ (a) as the input of UI-block1. (i) Results of Pro-UIGAN using ground-truth facial geometry priors. (j) Results of Pro-UInet (Pro-UIGAN trained without employing $L_{adv}^{l}$ and $L_{adv}^{g}$). (k) Results of Pro-UInet and Local-D. (l) Results of Pro-UIGAN.}
\label{figloss}
\end{figure}

To dissect the impacts of Global-D and Local-D, we provide the results of different Pro-UIGAN variants (Figs.~\ref{figloss}(j), (k), and (l)).
As shown in Fig.~\ref{figloss}(k), the Pro-UInet with Local-D generates visually pleasing details of facial components.
However, the global structure of the face is still blurry.
This is because it is difficult for the Local-D to impact the whole image during the backpropagation.
In contrast, as shown in Fig.~\ref{figloss}(l), our Pro-UIGAN captures not only local facial characteristics but also global profiles of faces.

\subsection{Objective Functions}
We utilize five loss terms for training our Pro-UIGAN, including an intensity similarity loss ($L_{mse}$), a feature similarity loss ($L_{f}$), a geometry similarity loss ($L_{h}$), a style loss ($L_{style}$), and a local-global discriminative loss ($L_{adv}^{l}$ and $L_{adv}^{g}$).
We illustrate the purpose of each introduced term in following subsections.

\subsubsection{Intensity similarity loss}
We constrain the hallucinated face to be close to the ground-truth in terms of image intensities.
Therefore, we employ an intensity similarity loss $L_{mse}$, expressed as:
\begin{equation}
\begin{aligned} L_{mse} &=\mathbb{E}_{\left(\hat{h}_{i}, h_{i}\right) \sim p(\hat{h}, h)}\left\|\hat{h}_{i}-h_{i}\right\|_{F}^{2}, \end{aligned}
\label{eqmse}
\end{equation}
where $p(\hat{h}, h)$ represents the joint distribution of the hallucinated face $\hat{h}_{i}$ and the corresponding ground-truth $h_{i}$.

\subsubsection{Feature similarity loss}
Since the intensity similarity loss $L_{mse}$ would lead to over-smoothed results~\cite{ledig2017photo} (see Fig.~\ref{figloss}(f)), we introduce a feature similarity loss~\cite{yu2019can} to improve our hallucinated faces.
The feature similarity loss $L_{f}$ measures the Euclidean distance between the high-level features of a hallucinated face and its ground-truth, written as:
\begin{equation}
\begin{aligned} L_{f} &=\mathbb{E}_{\left(\hat{h}_{i}, h_{i}\right) \sim p(\hat{h}, h)}\left\|\Phi\left(\hat{h}_{i}\right)-\Phi\left(h_{i}\right)\right\|_{F}^{2}, \end{aligned}
\label{eq3}
\end{equation}
where $\Phi(\cdot)$ represents the extracted feature maps of a layer in VGG-19~\cite{simonyan2014very}.
We use the layer ReLU32, which gives good empirical results in our experiments.
As seen in Fig.~\ref{figloss}(g), exploiting $L_{f}$ results in better hallucinated results with more photo-realistic facial details.

\subsubsection{Geometry similarity loss}
Aiming at maintaining the structural integrity of hallucinated faces as well as ensuring the accuracy of estimated facial geometry priors by CM-AMs, we employ a geometry similarity loss $L_{h}$~\cite{yu2018super} in training our CM-AMs, expressed as:
\begin{equation}
L_{h}=\mathbb{E}_{\left(f_{i}, h_{i}\right) \sim p(f, h)} \frac{1}{L} \sum_{l=1}^{L} \left\|{H}^{l}\left(f_{i}\right)-{H}^{FAN}\left(h_{i}\right)\right\|_{2}^{2},
\label{eq4}
\end{equation}
where ${H}^{l}\left(f_{i}\right)$ represents the estimated $l$-th facial landmark heatmap by our CM-AM on the intermediate feature maps ${f}^{i}$.
${H}^{FAN}\left(h_{i}\right)$ denotes the corresponding facial landmark heatmap generated by FAN~\cite{bulat2017far} on the ground-truth face image ${h}_{i}$.

\subsubsection{Style loss} 
Inspired by~\cite{liu2018image}, we introduce a style loss $L_{style}$ to enforce the style of the hallucinated face image to be similar to the ground-truth one.
The style loss $L_{style}$ is defined as:
\begin{equation}
\begin{aligned}
L_{\text {style}}=& \sum_{n=0}^{N-1}\left\|K_{n}\left(\phi_{n}\left(\hat{h}_{i}\right)^{T} \phi_{n}\left(\hat{h}_{i}\right)-\phi_{n}\left({h}_{i}\right)^{T} \phi_{n}\left({h}_{i}\right)\right)\right\|_{1},
\end{aligned}
\label{loss-smooth}
\end{equation}
where $\phi_{n}(\cdot)$ represents the extracted feature maps of the n-th layer in VGG-16~\cite{simonyan2014very}, and we employ the pool1, pool2, and pool3 layers.
$K_{n} = 1 /\left(C_{n} \cdot H_{n} \cdot W_{n}\right)$ is a normalization factor for the n-th VGG-16 layer. 
$C_{n}$, $H_{n}$, and $W_{n}$ are the channel number, height, and width of the extracted feature maps, respectively.

\subsubsection{Local-Global discriminative loss}
To generate visually appealing results, we incorporate class-specific discriminative information into our Pro-UInet by employing local-global discriminators, \emph{i.e.,} Local-D and Global-D.
Considering that the masked region shape may be irregular, we apply Local-D to the central facial regions, whose sizes are $16\times 16$ and $64\times 64$ pixels for the results of UI-block1 and UI-block3, respectively.
Our goal is to fool the local-global discriminators and make them fail to classify hallucinated images and real ones.

The objective function $L^{l}_{D}$ for the Local-D is defined as follows:
\begin{equation}
\begin{aligned}
L^{l}_{D} &=-\mathbb{E}_{\left(\hat{m}_{i}, m_{i}\right) \sim p(\hat{m}, m)}\left[\log {D}^{l}_{u}\left(m_{i}\right)+\log \left(1-{D}^{l}_{u}\left(\hat{m}_{i}\right)\right)\right],
\end{aligned}
\label{eq5}
\end{equation}
where $p(\hat{m}, m)$ represents the joint distribution of reconstructed central facial regions $\hat{m}_{i}$ and corresponding ground-truths $m_{i}$.
$D^{l}$ and $u$ represent the Local-D and its parameters.
To make the Local-D distinguish hallucinated facial regions from real ones, we minimize the loss $L^{l}_{D}$ and update its parameters $u$.

The objective function $L^{g}_{D}$ for the Global-D is defined as follows:
\begin{equation}
\begin{aligned} 
L^{g}_{D} &=-\mathbb{E}_{\left(\hat{h}_{i}, h_{i}\right) \sim p(\hat{h}, h)}\left[\log {D}^{g}_{v}\left(h_{i}\right)+\log \left(1-{D}^{g}_{v}\left(\hat{h}_{i}\right)\right)\right],
\end{aligned}
\label{eq6}
\end{equation}
where $D^{g}$ and $v$ represent the Global-D and its parameters.
To enable the Global-D to distinguish hallucinated faces from real ones, we minimize the loss $L^{g}_{D}$ and update its parameters $v$.

For the generator in our Pro-UInet, it aims to fool the local-global discriminators by producing realistic non-occluded HR faces.
Thus, the local discriminative loss $L^{l}_{adv}$ is written as:
\begin{equation}
\begin{aligned} L^{l}_{adv} &=-\mathbb{E}_{\hat{m}_{i} \sim p(\hat{m})} \log \left({D}^{l}_{u}\left(\hat{m}_{i}\right)\right). \end{aligned}
\label{eq7}
\end{equation}

Meanwhile, the global discriminative loss ${L}^{g}_{adv}$ is represented as:
\begin{equation}
\begin{aligned} {L}^{g}_{adv} &=-\mathbb{E}_{\hat{h}_{i} \sim p(\hat{h})} \log \left({D}^{g}_{v}\left(\hat{h}_{i}\right)\right). \end{aligned}
\label{eq8}
\end{equation}

In the network learning process, we minimize $L^{l}_{adv}$ and $L^{g}_{adv}$.

\subsection{Training Details}
Although UI-block1, UI-block2, and UI-block3 tackle the same hallucination subtask, their different inputs with diverse details attach their learning process with varying levels of difficulty.
Therefore, we adopt different loss terms for training different UI-blocks.

The objective function for the UI-block1, $L_{net1}$, is expressed as:
\begin{equation}
\begin{aligned} 
L_{net1}= L_{mse}^{a}+\alpha L_{f}^{a}+ L_{h}^{a}+ \gamma^{a} L_{style}^{a}+\psi {L}_{adv}^{l}.
\end{aligned}
\label{loss-UInet1}
\end{equation}

The objective function for the UI-block2, $L_{net2}$, is expressed as:
\begin{equation}
\begin{aligned} 
L_{net2}= L_{mse}^{b}+\alpha L_{f}^{b}+ L_{h}^{b}+\gamma^{b} L_{style}^{b}.
\end{aligned}
\label{loss-UInet2}
\end{equation}

The objective function for the UI-block3, $L_{net3}$, is expressed as:
\begin{equation}
\begin{aligned} 
 L_{net3}= L_{mse}^{c}+\alpha L_{f}^{c}+ L_{h}^{c}+\gamma^{c} L_{style}^{c}+\psi ({L}_{adv}^{l}+{L}_{adv}^{g}).
\end{aligned}
\label{loss-UInet3}
\end{equation}

Consequently, the total objective function of our Pro-UInet, $L_{G}$, is written as:
\begin{equation}
\begin{aligned}
 L_{G}= L_{net1}+ L_{net2}+ L_{net3}.
\end{aligned}
\label{loss-UIGAN}
\end{equation}

Since we aim to force hallucinated HR faces to be similar to real ones, and the feature similarity loss ($L_{f}$) and the discriminative loss ($L^{l}_{adv}$ and $L^{g}_{adv}$) are not used to measure the similarity between two images, we set lower weights on $L_{f}$, $L^{l}_{adv}$ and $L^{g}_{adv}$.
Thus, we set $\alpha$ and $\psi$ in Eqs.~\eqref{loss-UInet1}$-$\eqref{loss-UInet3} to 0.01.
Meanwhile, we set $\gamma^{a}, \gamma^{b}$, and $\gamma^{c}$ to 10, 10, and 1, respectively.

The training procedure of our Pro-UIGAN model includes three steps:
$\left(\romannumeral1\right)$ Pre-training the UI-block1 by $L_{net1}$ (Eq.~\eqref{loss-UInet1}) so as to initialize model parameter.
$\left(\romannumeral2\right)$ Pre-training the UI-block2 by $L_{net2}$ (Eq.~\eqref{loss-UInet2}). In this stage, the UI-block1 has been initialized.
$\left(\romannumeral3\right)$ Training the whole Pro-UIGAN model: Pro-UInet is trained by $L_{G}$ (Eq.~\eqref{loss-UIGAN}), and Local-D and Global-D are optimized by $L^{l}_{D}$ (Eq.~\eqref{eq5}) and $L^{g}_{D}$ (Eq.~\eqref{eq6}).
In step $\left(\romannumeral3\right)$, since our UI-block1 and UI-block2 have been initialized, we set the learning rates for training UI-block3, UI-block2, and UI-block1 to 10$^{-3}$, 10$^{-4}$, and 10$^{-4}$, respectively.

\section{Experiments}
\begin{figure*}[htb]
\centering
\vspace{-0.2cm}
\includegraphics[width=0.95\textwidth]{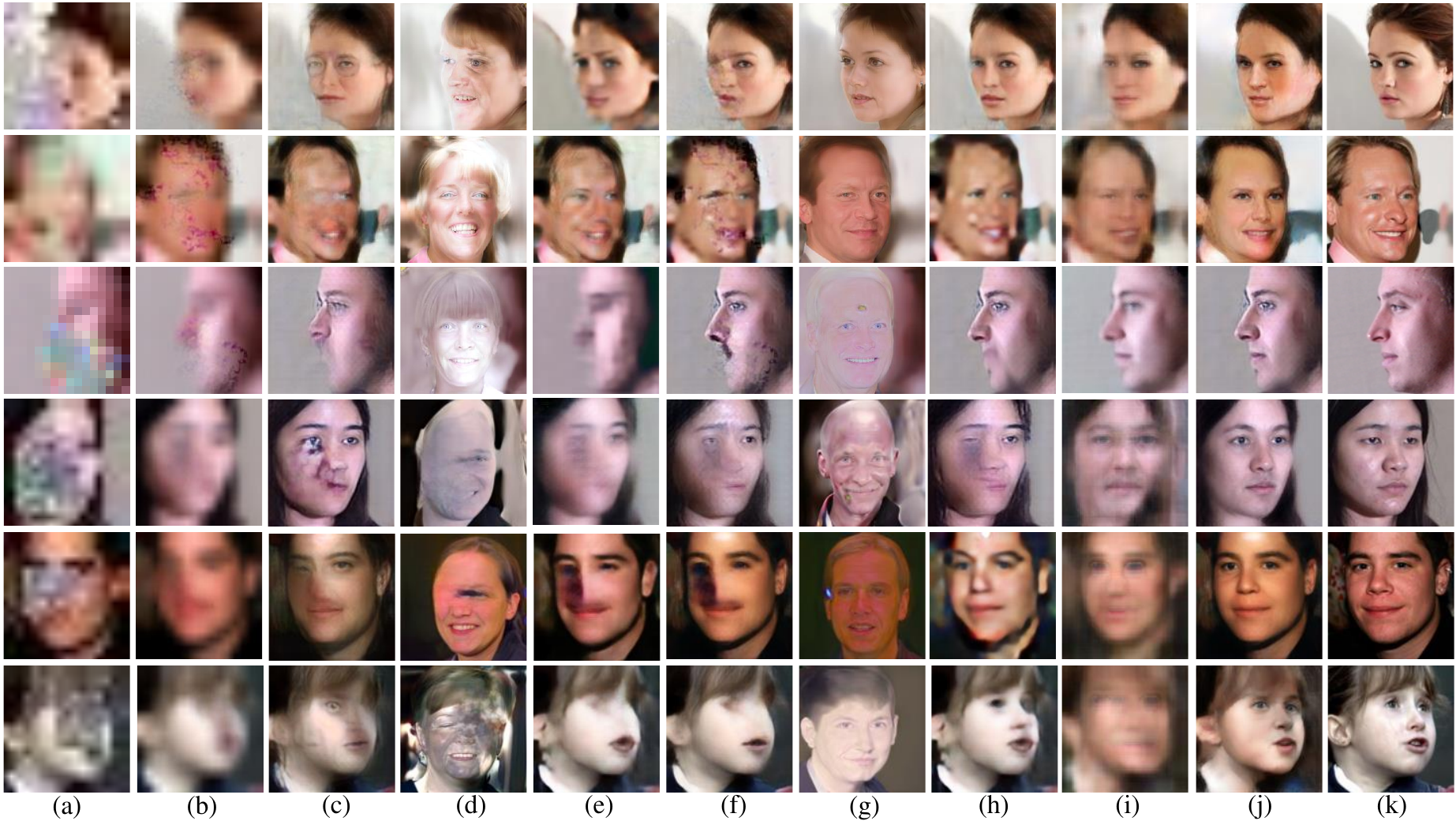}
\caption{Qualitative comparisons on the \textbf{CelebA-HQ}, \textbf{Multi-PIE}, and \textbf{Helen} databases. Columns: (a) Occluded LR faces ($16\times 16$ pixels). (b) Bicubic +~\cite{li2017generative}. (c)~\cite{yu2018face} +~\cite{li2017generative}. (d)~\cite{menon2020pulse} +~\cite{li2017generative}. (e)~\cite{zhang2021recursive} +~\cite{li2017generative}. (f)~\cite{li2017generative} +~\cite{yu2018face}. (g)~\cite{li2017generative} +~\cite{menon2020pulse}. (h)~\cite{li2017generative} +~\cite{zhang2021recursive}. (i)~\cite{cai2019fcsrj}. (j) Ours. (k) Ground-truths. The first two lines: testing samples from {\bf CelebA-HQ}. The middle two lines: testing samples from {\bf Multi-PIE}. The last two lines: testing samples from {\bf Helen}.}
\label{viewcompare}
\end{figure*}

\begin{table*}[t]
\caption{Quantitative comparisons on the \textbf{CelebA-HQ}, \textbf{Multi-PIE}, and \textbf{Helen} databases.}
\centering
\begin{tabular}{@{}l|l|cccccc|cccccc@{}}
\toprule
\multirow{3}{*}{SR Method} & \multirow{3}{*}{Mask Type} & \multicolumn{2}{c|}{CelebA-HQ}    & \multicolumn{2}{c|}{Multi-PIE}                                                             & \multicolumn{2}{c|}{Helen}                                                              
& \multicolumn{2}{c|}{CelebA-HQ}                                                             & \multicolumn{2}{c|}{Multi-PIE}                                                             & \multicolumn{2}{c}{Helen}                                                               
\\ \cline{3-14} &              
& \multicolumn{6}{c|}{SR+PA~\cite{li2017generative}}                                                                 & \multicolumn{6}{c}{PA~\cite{li2017generative}+SR}                                                                  \\ \cline{3-14} &    &     \multicolumn{1}{c|}{PSNR}                             & \multicolumn{1}{c|}{SSIM}                            & \multicolumn{1}{c|}{PSNR}                             & \multicolumn{1}{c|}{SSIM}                            & \multicolumn{1}{c|}{PSNR}                             & \multicolumn{1}{c|}{SSIM}                            & \multicolumn{1}{c|}{PSNR}                             & \multicolumn{1}{c|}{SSIM}                            & \multicolumn{1}{c|}{PSNR}                             & \multicolumn{1}{c|}{SSIM}                            & \multicolumn{1}{c|}{PSNR}                             & \multicolumn{1}{c}{SSIM}                            \\ \cline{1-14} 
\multirow{2}{*}{Bicubic}   & Irregular                    & \multicolumn{1}{c|}{12.926}                           & \multicolumn{1}{c|}{0.371}                           & \multicolumn{1}{c|}{11.702}                           & \multicolumn{1}{c|}{0.338}                           & \multicolumn{1}{c|}{12.587}                           & \multicolumn{1}{c|}{0.364}                           & \multicolumn{1}{c|}{14.078}                           & \multicolumn{1}{c|}{0.419}                           & \multicolumn{1}{c|}{13.109}                           & \multicolumn{1}{c|}{0.375}                           & \multicolumn{1}{c|}{12.630}                           & \multicolumn{1}{c}{0.368}                           \\
                           & Square                     & \multicolumn{1}{c|}{14.038}                           & \multicolumn{1}{c|}{0.411}                           & \multicolumn{1}{c|}{12.761}                           & \multicolumn{1}{c|}{0.370}                           & \multicolumn{1}{c|}{13.665}                           & \multicolumn{1}{c|}{0.389}                           & \multicolumn{1}{c|}{15.322}                           & \multicolumn{1}{c|}{0.457}                           & \multicolumn{1}{c|}{14.215}                           & \multicolumn{1}{c|}{0.416}                           & \multicolumn{1}{c|}{13.763}                           & \multicolumn{1}{c}{0.392}                           \\ \cline{1-14} 
\multirow{2}{*}{FHC~\cite{yu2018face}}       & Irregular                    & \multicolumn{1}{c|}{16.575}                           & \multicolumn{1}{c|}{0.490}                           & \multicolumn{1}{c|}{15.727}                           & \multicolumn{1}{c|}{0.464}                           & \multicolumn{1}{c|}{15.288}                           & 0.458                           & \multicolumn{1}{c|}{17.953}                           & \multicolumn{1}{c|}{0.549}                           & \multicolumn{1}{c|}{16.704}                           & \multicolumn{1}{c|}{0.491}                           & \multicolumn{1}{c|}{16.241}                           & \multicolumn{1}{c}{0.480}                           \\
                           & Square                                      & \multicolumn{1}{c|}{16.950}                           & \multicolumn{1}{c|}{0.503}                           & \multicolumn{1}{c|}{16.448}                           & \multicolumn{1}{c|}{0.482}                           & \multicolumn{1}{c|}{16.003}                           & 0.476                           & \multicolumn{1}{c|}{19.224}                           & \multicolumn{1}{c|}{0.609}                           & \multicolumn{1}{c|}{18.198}                           & \multicolumn{1}{c|}{0.573}                           & \multicolumn{1}{c|}{17.715}                           & \multicolumn{1}{c}{0.546}                           \\ \cline{1-14} 
\multirow{2}{*}{PULSE~\cite{menon2020pulse}}     & Irregular                    & \multicolumn{1}{c|}{12.219}                           & \multicolumn{1}{c|}{0.365}                           & \multicolumn{1}{c|}{7.938}                            & \multicolumn{1}{c|}{0.169}                           & \multicolumn{1}{c|}{9.794}                           & \multicolumn{1}{c|}{0.235}                           & \multicolumn{1}{c|}{13.992}                           & \multicolumn{1}{c|}{0.411}                           & \multicolumn{1}{c|}{8.008}                            & \multicolumn{1}{c|}{0.173}                           & \multicolumn{1}{c|}{9.976}                           & \multicolumn{1}{c}{0.238}                           \\
                           & Square               & \multicolumn{1}{c|}{13.106}                           & \multicolumn{1}{c|}{0.375}                           & \multicolumn{1}{c|}{8.451}                            & \multicolumn{1}{c|}{0.182}                           & \multicolumn{1}{c|}{10.516}                           & 0.273                           & \multicolumn{1}{c|}{15.270}                           & \multicolumn{1}{c|}{0.456}                           & \multicolumn{1}{c|}{9.237}                            & \multicolumn{1}{c|}{0.206}                           & \multicolumn{1}{c|}{11.128}                           & \multicolumn{1}{c}{0.294}                           \\\cline{1-14} 
\multirow{2}{*}{Re-CPGAN~\cite{zhang2021recursive}}    & Irregular                    & \multicolumn{1}{c|}{17.962}                           & \multicolumn{1}{c|}{0.548}                           & \multicolumn{1}{c|}{15.973}                           & \multicolumn{1}{c|}{0.467}                           & \multicolumn{1}{c|}{16.689}                           & \multicolumn{1}{c|}{0.490}                           & \multicolumn{1}{c|}{18.754}                           & \multicolumn{1}{c|}{0.598}                           & \multicolumn{1}{c|}{16.974}                           & \multicolumn{1}{c|}{0.497}                           & \multicolumn{1}{c|}{17.075}                           & \multicolumn{1}{c}{0.508}                           \\
                           & Square                          & \multicolumn{1}{c|}{18.276}                           & \multicolumn{1}{c|}{0.575}                           & \multicolumn{1}{c|}{16.918}                           & \multicolumn{1}{c|}{0.501}                           & \multicolumn{1}{c|}{17.231}                           & 0.536                           & \multicolumn{1}{c|}{19.979}                           & \multicolumn{1}{c|}{0.611}                           & \multicolumn{1}{c|}{18.122}                           & \multicolumn{1}{c|}{0.568}                           & \multicolumn{1}{c|}{18.893}                           & \multicolumn{1}{c}{0.602}                           \\  \cline{1-14} 
\multirow{2}{*}{FCSR-GAN~\cite{cai2019fcsrj}}  & Irregular                    & \multicolumn{1}{c|}{20.088}                           & \multicolumn{1}{c|}{0.617}                           & \multicolumn{1}{c|}{19.963}                           & \multicolumn{1}{c|}{0.610}                           & \multicolumn{1}{c|}{18.872}                           & 0.601                           & \multicolumn{1}{c|}{20.088}                           & \multicolumn{1}{c|}{0.617}                           & \multicolumn{1}{c|}{19.963}                           & \multicolumn{1}{c|}{0.610}                           & \multicolumn{1}{c|}{18.872}                           & \multicolumn{1}{c}{0.601}                           \\
                           & Square                 & \multicolumn{1}{c|}{23.010}                           & \multicolumn{1}{c|}{0.698}                           & \multicolumn{1}{c|}{21.327}                           & \multicolumn{1}{c|}{0.649}                           & \multicolumn{1}{c|}{20.745}                           & 0.627                           & \multicolumn{1}{c|}{23.010}                           & \multicolumn{1}{c|}{0.698}                           & \multicolumn{1}{c|}{21.327}                           & \multicolumn{1}{c|}{0.649}                           & \multicolumn{1}{c|}{20.745}                           & \multicolumn{1}{c}{0.627}                           \\ \cline{1-14} 
\multirow{2}{*}{Pro-UIGAN} & Irregular       & \multicolumn{1}{c|}{24.534}                           & \multicolumn{1}{c|}{0.706}                           & \multicolumn{1}{c|}{22.119}                           & \multicolumn{1}{c|}{0.668}                           & \multicolumn{1}{c|}{21.522}                           & \multicolumn{1}{c|}{0.652}                        & \multicolumn{1}{c|}{24.534}                           & \multicolumn{1}{c|}{0.706}                           & \multicolumn{1}{c|}{22.119}                           & \multicolumn{1}{c|}{0.668}                           & \multicolumn{1}{c|}{21.522}                           & \multicolumn{1}{c}{0.652}                      \\
                           & Square                      & \multicolumn{1}{c|}{\textbf{25.682}}                           & \multicolumn{1}{c|}{\textbf{0.770}}                           & \multicolumn{1}{c|}{\textbf{23.505}}                           & \multicolumn{1}{c|}{\textbf{0.724}}                           & \multicolumn{1}{c|}{\textbf{22.459}}                           & \multicolumn{1}{c|}{\textbf{0.681}}                        & \multicolumn{1}{c|}{\textbf{25.682}}                           & \multicolumn{1}{c|}{\textbf{0.770}}                           & \multicolumn{1}{c|}{\textbf{23.505}}                           & \multicolumn{1}{c|}{\textbf{0.724}}                           & \multicolumn{1}{c|}{\textbf{22.459}}                           & \multicolumn{1}{c}{\textbf{0.681}}                      \\
\bottomrule
\end{tabular}
\label{table1}
\end{table*}

\subsection{Experimental Setup}
\subsubsection{Databases}
Our Pro-UIGAN is trained and tested on popular face databases, \emph{i.e.,} the CelebA-HQ database~\cite{liu2015faceattributes}, the Multi-PIE database~\cite{gross2010multi}, and the Helen database~\cite{le2012interactive}.

\textbf{Multi-PIE}~\cite{gross2010multi} provides 750K+ images of 337 individuals under different conditions.
We select $12,912$ images of all the individuals spanning across various poses (0$^{o}$, $\pm$15$^{\circ}$, $\pm$30$^{\circ}$, $\pm$45$^{\circ}$, $\pm$60$^{\circ}$, $\pm$75$^{\circ}$, $\pm$90$^{\circ}$) as well as expressions (``squint”, ``disgust”, ``neutral”, ``smile”, ``surprise”, and ``scream”).
Our training set contains $12,000$ images belonging to the former 250 individuals, and our testing set includes 912 images belonging to the remaining 87 individuals.

\textbf{CelebA-HQ}~\cite{karras2017progressive} consists of $30,000$ HR face images under various poses, expressions, and backgrounds.
Each face image has a binary segmentation mask as well as 19 labeled facial attributes, \emph{e.g.,} eyes, mouth, hat, neck, skin, etc.
We employ the standard split for CelebA-HQ in our experiments, where $28,432$ images are for training, and $1,568$ images are for testing.

\textbf{Helen}~\cite{le2012interactive} is composed of $2,330$ in-the-wild face images with labeled facial components, \emph{e.g.,} eyebrows, lips, nose, skin, hairs, etc.
We use $2,280$ images to construct the training set and the remaining 50 images to form the testing set.
Specifically, we conduct data augmentation for the training set.
We rotate all the face images by $\pm$90$^{\circ}$, $\pm$180$^{\circ}$, $\pm$270$^{\circ}$, and then flip them horizontally.
As a result, we augment seven additional images for each original one.

\subsubsection{Implementation details}
First, we detect the faces in all the databases by the Dlib face detector, align them to the upright position, crop main facial areas from original images, and resize them to $128\times 128$ pixels with bilinear interpolation.
Those processed images form our ground-truth HR images without occlusions.
Then, we generate both irregular masks following PConv~\cite{liu2018image} and square masks with different sizes randomly ranging from $16\times 16$ to $64\times 64$ pixels to construct the occluded HR faces.
Afterward, we generate the occluded LR faces ($16\times 16$ pixels) by downsampling the occluded HR ones with bilinear interpolation.
As a result, we construct occluded LR$/$non-occluded HR face pairs for each database.
We conduct our experiments with the PyTorch framework on an NVIDIA Titan 2080 GPU card.
The Pro-UIGAN model takes about five days to converge.
Our large-scale occluded LR$/$non-occluded HR face pair datasets and the code will be available on \url{https://github.com/SEU-yang.}

\subsubsection{Competing methods}
We conduct comparative experiments in the following three manners:
 \begin{itemize}
     \item SR+PA: face SR methods (FSRnet~\cite{chen2018fsrnet}, FHC~\cite{yu2018face}, or PULSE~\cite{menon2020pulse}) followed by face inpainting techniques (GFC~\cite{li2017generative});
      \item PA+SR: face inpainting techniques (GFC~\cite{li2017generative}) followed by face SR methods (FSRnet~\cite{chen2018fsrnet}, FHC~\cite{yu2018face}, or PULSE~\cite{menon2020pulse}) (bicubic interpolation is used to resize images);
      \item Joint manner: FCSR-GAN~\cite{cai2019fcsrj} and our Pro-UIGAN.
 \end{itemize}

In the first manner (SR+PA), we super-resolve the occluded LR faces first and then inpaint the upsampled results.
In the second manner (PA+SR), we first inpaint occluded LR faces and then super-resolve the inpainted faces.
In the third manner (Joint manner), both FCSR-GAN~\cite{cai2019fcsrj} and Pro-UIGAN handle face SR and face inpainting in a unified framework.
We retrain all these prior methods on the utilized datasets for fair comparisons.

\subsection{Qualitative Evaluation}

\begin{table}[t]
\caption{Ablation study of facial geometry priors.}
\centering
\begin{tabular}{@{}cccccc@{}}
\bottomrule
\multirow{2}{*}{\begin{tabular}[c]{@{}c@{}} \end{tabular}} & \multicolumn{2}{c}{CelebA-HQ}    & \multicolumn{2}{c}{Multi-PIE}       \\ \cline{2-5} 
                                                                        & PSNR            & SSIM           & PSNR            & SSIM           \\\bottomrule 
P-FP                                  &    22.866          & 0.683                          &  20.341          & 0.647           \\
 P+GT                                     & \textbf{26.448}          & \textbf{0.797}                           & \textbf{24.236}         & \textbf{0.743}             \\ 
Pro-UIGAN                                                                 & 25.682      & 0.770          & 23.505         & 0.724  \\ \bottomrule
\end{tabular}
\label{table-abm}
\end{table}

\begin{table}[t]
\caption{Ablation study of different training losses.}
\centering
\begin{tabular}{@{}c|c|c|c|cc@{}}
\toprule
\multirow{2}{*}{} & \multicolumn{2}{c|}{CelebA-HQ} & \multicolumn{2}{c}{Multi-PIE} \\ \cmidrule(l){2-5} 
                                  & PSNR           & SSIM         & PSNR         & SSIM        \\ \midrule
$L_{G}^{-}$                      & 19.136       & 0.605            & 18.473         & 0.579      \\
$L_{G}^{\dagger}$                & 20.069       & 0.614              & 18.990         & 0.593      \\
$L_{G}^{\ddagger}$              & 20.872       & 0.628              & 19.281         & 0.610       \\
$L_{G}^{\star}$           & 24.960       & 0.756            & 22.755         & 0.674        \\
$L_{G}$            & \textbf{25.682}      & \textbf{0.770}          & \textbf{23.505}         & \textbf{0.724}       \\ \bottomrule
\end{tabular}
\label{table2}
\end{table}

\begin{figure}[t]
\centering
\includegraphics[height=3.2cm,width=0.48\textwidth]{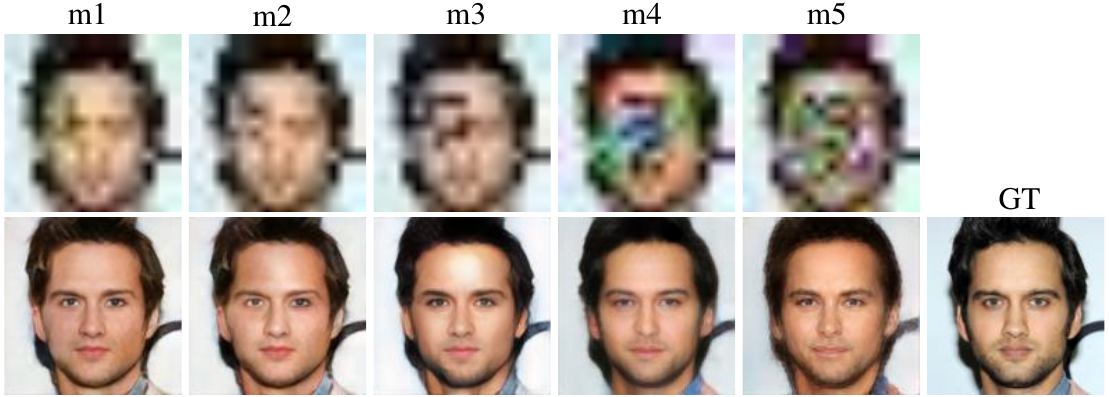}
\caption{Qualitative evaluation on different mask sizes.}
\label{fig-masksize}
\end{figure}

\begin{figure}[t]
\centering
\begin{minipage}[b]{.24\textwidth}
\centering
\includegraphics[height=3cm]{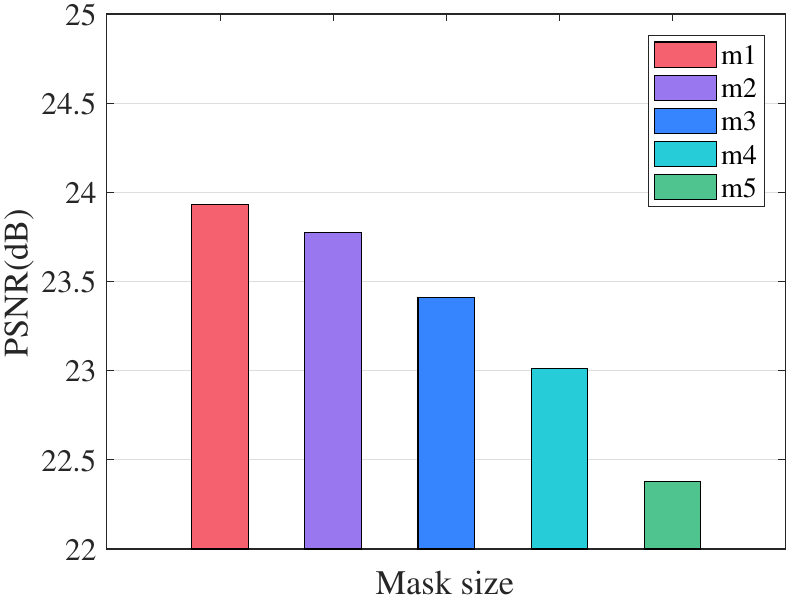}
\subcaption{Multi-PIE}
\end{minipage}
\begin{minipage}[b]{.24\textwidth}
\centering
\includegraphics[height=3cm]{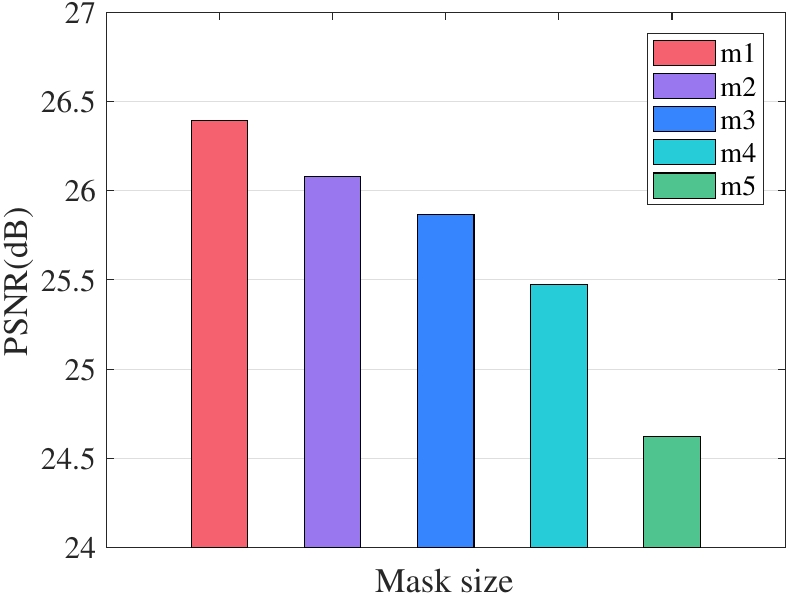}
\subcaption{CelebA-HQ}
\end{minipage}
\caption{Quantitative evaluation on different mask sizes.}
\label{fig-masksize-psnr}
\end{figure}

Fig.~\ref{viewcompare} illustrates the qualitative results of the compared methods.
As shown in Fig.~\ref{viewcompare}(b), different combinations of bicubic interpolation and face inpainting methods~\cite{li2017generative} fail to reconstruct authentic facial details.
Because bicubic interpolation generates new pixels from neighboring ones via a simple interpolation process, it leads the upsampled HR images to lack details.
Without enough face details in the given HR images, the face inpainting method fails to segment facial areas accurately.
Consequently, it generates erroneous results, such as over-smoothed facial details and distorted contours.
Similarly, the results of SR+PA and PA+SR methods suffer from corrupted facial regions and ghosting artifacts (see Figs.~\ref{viewcompare}(c)-(h)).

Since FCSR-GAN~\cite{cai2019fcsrj} super-resolves and inpaints occluded LR faces via an integrated framework, it generates better results compared to previous works~\cite{chen2018fsrnet,yu2018face,menon2020pulse,li2017generative} treating the two tasks independently and sequentially.
However, as a ``one-shot deal” method, FCSR-GAN does not possess a ``looking back” ability to refine the generated results with flaws, which might occur in those difficult input samples.
For example, when the input LR faces are under large poses or complex expressions (see Fig.~\ref{viewcompare}(a)), FCSR-GAN produces blurry HR faces, which are shown in Fig.~\ref{viewcompare}(i).

As shown in Fig.~\ref{viewcompare}(j), our Pro-UIGAN generates visually appealing non-occluded HR faces.
To illustrate the superiority of our method, we show challenging cases with extreme poses and expressions (\emph{e.g.,} the first, third, and sixth lines in Fig.~\ref{viewcompare}(a)).
We can find that the results generated by Pro-UIGAN are more authentic and vivid.
The reason is that Pro-UIGAN exploits a multi-stage progressive hallucination strategy, refining the hallucinated faces in a multi-stage and joint manner.

\begin{figure*}[t]
\centering
\includegraphics[height=4cm,width=0.95\textwidth]{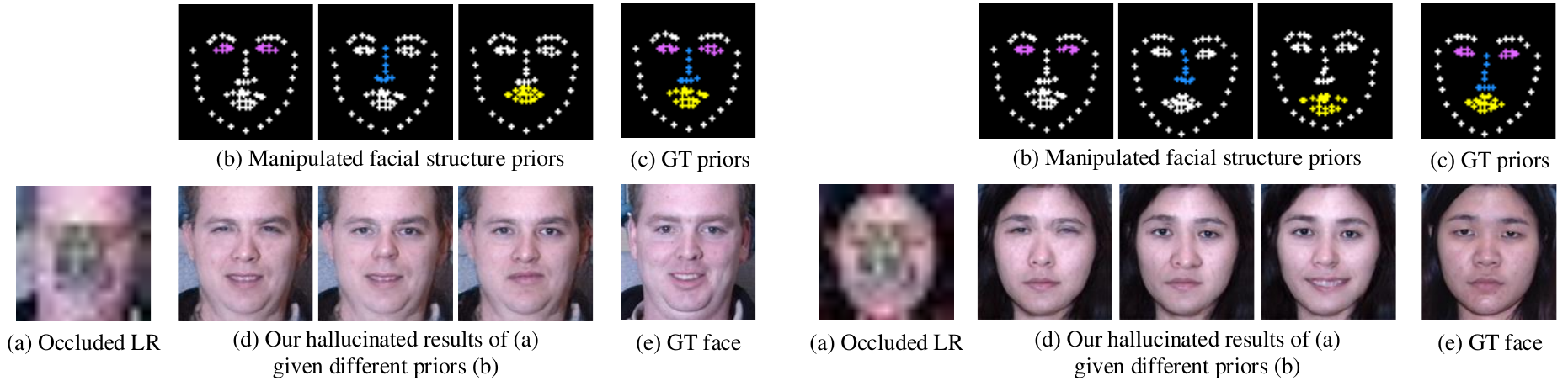}
\caption{Facial attribute editing results.}
\label{fig-geometry}
\end{figure*}

\subsection{Quantitative Evaluation}
To evaluate the hallucination performance quantitatively, we calculate the average Peak Single-to-Noise Ratio (PSNR) as well as Structural Similarity (SSIM) values of all methods and report them in Tab.~\ref{table1}.

Based on the reported results in Tab.~\ref{table1}, we can find that our Pro-UIGAN obtains the best quantitative results on all the databases.
For example, on the CelebA-HQ testing set with square masks, Pro-UIGAN surpasses the other baselines by improving the PSNR by $11.6\%$.
Other than that, when given occluded images with irregular mask shapes, our method still outperforms other methods, \emph{e.g.,} 4.446 higher than FCSR-GAN in PSNR.
Since our model is accomplished by the multi-stage progressive hallucination strategy, it inpaints occluded face images progressively.

Furthermore, Tab.~\ref{table1} also indicates that inpainting followed by super-resolution leads to slightly higher quantitative results than super-resolution followed by inpainting.
This implies that super-resolving non-occluded LR facial patterns are easier than occluded ones.

\subsection{Ablation Analysis}
\subsubsection{Impacts of facial geometry priors}
In our work, we exploit facial geometry priors, \emph{i.e.,} facial landmark heatmaps, for joint face SR and inpainting.
We provide a discussion about how many improvements the facial geometry priors bring.

We formulate two comparing network structures below (we denote Pro-UIGAN as P, estimated facial priors as FP, and ground-truth facial priors as GT): 
\begin{itemize}
\item P-FP:  we remove the prior estimation network, \textit{i.e.}, CM-AM. Here, we use three residual blocks~\cite{he2016deep} to replace the CM-AM in each UI-block, keeping the model size consistent.
\item P+GT: we use ground-truth facial landmark heatmaps instead of estimated facial priors.
\end{itemize}

As illustrated in Fig.~\ref{figloss}, the result of P+GT (see Fig.~\ref{figloss}(i)) shows more accurate facial geometry than the results of P-FP (see Fig.~\ref{figloss}(h)) and our Pro-UIGAN (see Fig.~\ref{figloss}(l)).
This indicates the importance of facial geometry priors in the whole process: accurate facial prior knowledge significantly reduces the ambiguous mapping caused by occlusions and thus facilitates the upsampling procedures.
As shown in Tab.~\ref{table-abm}, P+GT (with the ground-truth facial geometry priors) outperforms Pro-UIGAN (with the estimated facial geometry prior) and P-FP (without prior information) with the PSNR improvement of 0.766 dB and 4.431 dB on CelebA-HQ, respectively.
These results demonstrate the efficacy of the proposed CM-AM.

\subsubsection{Impacts of loss terms}
We provide the results of our Pro-UIGAN trained by using different losses on Multi-PIE and CelebA-HQ (see Tab.~\ref{table2} and Fig.~\ref{figloss}).
We denote the compared Pro-UIGAN variants as follows: $\left(\romannumeral1\right)$ $L_{G}^{-}$: $L_{mse}$ and $L_{h}$; $\left(\romannumeral2\right)$ $L_{G}^{\dagger}$: $L_{mse}$, $L_{f}$ only for UI-block2 and UI-block3, and $L_{h}$;
$\left(\romannumeral3\right)$ $L_{G}^{\ddagger}$: $L_{mse}$, $L_{f}$, and $L_{h}$; 
$\left(\romannumeral4\right)$ $L_{G}^{\star}$: $L_{mse}$, $L_{f}$, $L_{h}$, $L_{style}$, and $L^{l}_{adv}$;
$\left(\romannumeral5\right)$ $L_{G}$: $L_{mse}$, $L_{f}$, $L_{h}$, $L_{style}$, $L^{l}_{adv}$, and $L^{g}_{adv}$. 
Note that $L_{h}$ is a prerequisite constraint in training our CM-AMs.

Fig.~\ref{figloss}(f) shows that only exploiting the intensity similarity loss $L_{mse}$ results in overly smooth results.
Therefore, we introduce a feature similarity loss $L_{f}$ to improve the visual quality (see Fig.~\ref{figloss}(f)).
We provide the quantitative results (\emph{i.e.}, $L_{G}^{-}$, $L_{G}^{\dagger}$ and $L_{G}^{\ddagger}$ in Tab.~\ref{table2}) to confirm the reasonable of visual results.
Meanwhile, we demonstrate that $L_{f}$ for UI-block1 is able to reduce the reconstruction errors of coarsely hallucinated faces in the first UI-block rather than spreading the errors through the entire Pro-UInet, and thus improves the model performance (see $L_{G}^{\dagger}$ and $L_{G}^{\ddagger}$ in Tab.~\ref{table2}).
Furthermore, as indicated in Tab.~\ref{table2} ($L_{G}^{\star}$) and Fig.~\ref{figloss}(k), the Local-D improves the model performance.
However, the face in Fig.~\ref{figloss}(k) still suffers from global structural inconsistency.
Finally, after incorporating Global-D, Pro-UIGAN achieves the best results (see Fig.~\ref{figloss}(l)) with the highest quantitative scores ($L_{G}$ in Tab.~\ref{table2}).

\begin{table}[t]
\centering
\caption{Efficiency comparisons on the CelebA-HQ testing set.}
\begin{tabular}{@{}c|c|c|c|c@{}}
\toprule
\multicolumn{2}{c|}{Method}                                                         & \begin{tabular}[c]{@{}c@{}}Model\\ Size\\ (KB)\end{tabular} & \begin{tabular}[c]{@{}c@{}}Running\\ Time\\ (ms)\end{tabular} & PSNR (dB)     \\ \midrule
\multirow{3}{*}{\begin{tabular}[c]{@{}c@{}}PA+SR \\ SR+PA\end{tabular}} 
                                                                      & FHC~\cite{yu2018face}         & 71,705                                                           & 289.12                                                               & 19.224/16.950   \\ 
                                                                      & PULSE~\cite{menon2020pulse}      & 118,522    &  5128.36                                                                                                                       & 15.270/13.106  \\& Re-CPGAN~\cite{zhang2021recursive}        & 98,946                                                           & 372.14                                                               & 19.979/18.276 \\ \midrule
\multirow{2}{*}{\begin{tabular}[c]{@{}c@{}}Joint\\ manner\end{tabular}} & FCSR-GAN~\cite{cai2019fcsrj}         & 112,248                                                          & 145.69                                                               & 23.010        \\  
                                                                      & Pro-UIGAN     & \textbf{41,529}                                                            & \textbf{34.21}                                                               & \textbf{25.682}        \\ \bottomrule
\end{tabular}
\label{efficiency}
\end{table}

\begin{figure}[t]
\centering
\includegraphics[height=17cm,width=0.48\textwidth]{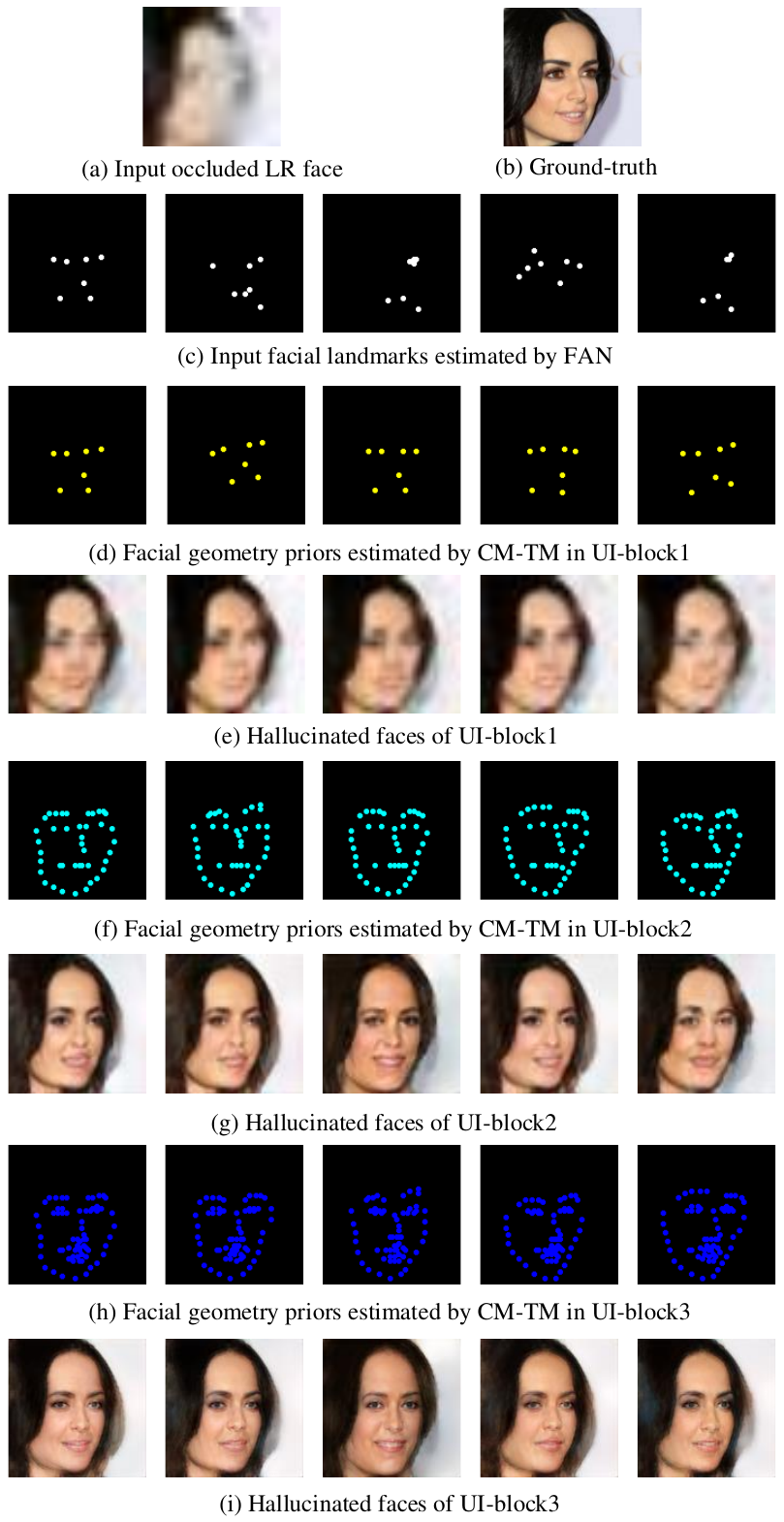}
\caption{Analysis of our multi-stage progressive hallucination strategy. Zoom in to see details.}
\label{fig-lanmarks}
\end{figure}

\subsection{Impacts of Regular and Irregular Masks}
\textbf{Regular Masks.} We evaluate the impacts of different mask sizes on our model.
The mask sizes are as follows: (a) m1: $16\times 16$ pixels, (b) m2: $24\times 24$ pixels, (c) m3: $32\times 32$ pixels, (d) m4: $48\times 48$ pixels, (e) m5: $64\times 64$ pixels.
Figs.~\ref{fig-masksize} and~\ref{fig-masksize-psnr} show the qualitative and quantitative results.
Although the performance gradually drops when the mask size increases, our Pro-UIGAN performs well for all mask sizes, even when the size is in a large value (\emph{i.e.,} m5).

\textbf{Irregular Masks.} We demonstrate that our Pro-UIGAN can also handle irregular masks, which is challenging for state-of-the-art face inpainting methods~\cite{li2017generative,2018Geometry,liu2017semantically}.
As shown in Fig.~\ref{viewcompare}, our model achieves more photo-realistic visual results on the LR faces with irregular occlusions.
Tab.~\uppercase\expandafter{\romannumeral1} also demonstrates that our Pro-UIGAN outperforms the state-of-the-arts with a large margin on quantitative performance.

\subsection{Facial Attributes Manipulation}
After hallucinating the occluded LR faces, users may not be satisfied with the generated facial attributes and want to manipulate them.
We demonstrate that our model allows the users to conduct interactive facial attribute editing on the hallucinated face.
As shown in Fig.~\ref{fig-geometry}, we can change the facial geometry priors to modify one attribute (\emph{i.e.,} mouth, nose, or eyes) while other attributes keep similar to the ground-truth.

\subsection{Efficiency Analysis}
We provide efficiency comparisons for our Pro-UIGAN and the competing baselines in Tab.~\ref{efficiency}.
Compared with the prior methods, our Pro-UIGAN requires the smallest running time (34.21 ms) and is more lightweight, which indicates the potential capability to apply Pro-UIGAN in low-resource scenarios.

\begin{figure}[t]
\centering
\includegraphics[height=4cm,width=0.48\textwidth]{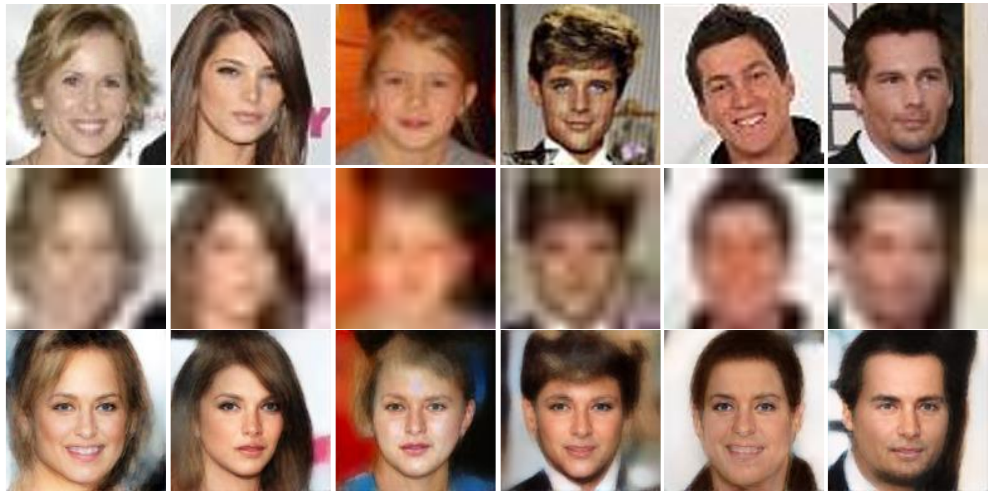}
\caption{Results on real-life unaligned LR face images. First row: real-life unaligned LR faces. Second row: input aligned LR faces ($16\times 16$ pixels). Third row: the results of our model.}
\label{fig-real-alignment}
\end{figure}

\subsection{Analysis of Our Progressive Hallucination Strategy}
\textbf{Why Pro-UIGAN works?}~\cite{karras2018progressive,morerio2017curriculum} have proved that the ``start easy” paradigm will likely guide the learning process of neural networks.
In our work, we propose to progressively hallucinate the occluded LR faces under the guidance of estimated facial geometry priors (\emph{i.e.,} facial landmark heatmaps).
As the network grows, it estimates more accurate facial geometry priors and reconstructs more high-quality faces.
Moreover, by initializing the parameter of the former UI-blocks step-by-step, we guarantee the convergence of our model during training.
In Fig.~\ref{fig-lanmarks}, we provide the estimated facial geometry priors and the hallucinated results by our multi-stage networks.
We can observe that the hallucination process is consistent with our hypothesis: the first UI-block roughs out the missing contents and generates a coarse inpainted and upsampled face, then the subsequent UI-blocks further refine the coarse hallucinated face, generating finer hallucinated ones.

\textbf{Robustness towards inaccurate input landmarks.}
Inspired by our multi-stage progressive hallucination strategy, our Pro-UInet stacks a series of UI-blocks.
Each UI-block comprises a CM-AM and a TUM.
Our CM-AM is designed to learn facial geometry priors from the input facial and landmark features under the supervision of geometry similarity loss $L_{h}$.
$L_{h}$ constrains the estimated facial geometry priors to be close to those of the ground-truth HR face.
Moreover, we pretrain the former UI-blocks so as to initialize model parameters and Sec.~\uppercase\expandafter{\romannumeral3}. G provides the details.
In this manner, we can guarantee the performance of CM-AM in the first UI-block.
Fig.~\ref{fig-lanmarks}(d) shows the estimated landmarks by CM-AM in the first UI-block.
The final results demonstrate that our CM-AM provides effective facial geometry priors even when input landmarks have errors (see Fig.~\ref{fig-lanmarks}(i)).

\begin{figure}[t]
\centering
\includegraphics[height=9cm,width=0.45\textwidth]{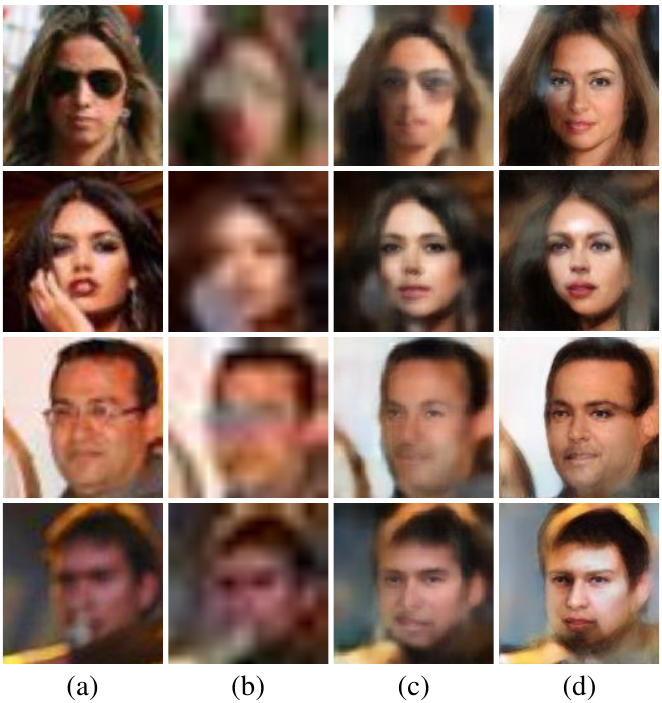}
\caption{Face hallucination results of real-life occluded LR faces. (a) Real-life occluded LR faces. (b) Input masked LR faces ($16\times 16$ pixels). (c)~\cite{cai2019fcsrj}. (d) Ours.}
\label{fig-mask}
\end{figure}

\subsection{Performance on Real-life Unaligned LR Faces}
Our model can also effectively handle real-life unaligned LR faces.
We select the non-occluded faces smaller than $48\times 48$ pixels from the CelebA database~\cite{liu2015faceattributes} for testing.
Here, we use MTCNN~\cite{zhang2016joint} to align the unaligned LR faces first.
As illustrated in Fig.~\ref{fig-real-alignment}, our Pro-UIGAN can generate authentic facial details when given unaligned LR face images.

\subsection{Performance on Real-life Occluded LR Faces}
We evaluate the performance of our Pro-UIGAN on hallucinating real-life occluded LR faces.
We select some naturally occluded LR face images from the CelebA database~\cite{liu2015faceattributes}.
Since such regions in the images are occluded or not is subjective, we give users this option to appoint the occluded regions through appending masks.
Then, we obtain the input masked LR faces by downsampling the processed images.
The results in Fig.~\ref{fig-mask} show that our model can hallucinate coarse facial appearances and remove real occlusions.
However, ghosting artifacts appear on the hallucinated faces.

This deterioration is mainly because there exist significant domain shifts between real-life occluded LR faces and our training data.
Moreover, the domain shifts would cause errors in both face SR and inpainting procedures.
Our future research will address the real-world degradation factors.

\subsection{Cross-dataset Validation}
The cross-dataset validation is used to evaluate the generalization ability of our Pro-UIGAN model.
We train our model on one dataset and test it on another, \emph{i.e.,} training on CelebA-HQ but testing on Multi-PIE.
As shown in Tab.~\ref{table-cross}, compared with the intra-database testing results, the cross-dataset validation results look relatively unsatisfied.
In our future work, we will explore efficient domain adaption methods for cross-domain face hallucination.

\begin{table}[t]
\caption{Performance of cross-dataset verification.}
\centering
\begin{threeparttable}
\begin{tabular}{@{}c|c|c|c@{}}
\toprule
Model              & CelebAHQ     & Multi-PIE    & Helen        \\ \midrule
Pro-UIGAN-C & \textbf{25.562/0.770} & 20.336/0.620 & 21.009/0.631 \\
Pro-UIGAN-M & 19.238/0.610 & \textbf{23.505/0.724} & 18.840/0.598 \\
Pro-UIGAN-H    & 20.723/0.624 & 17.947/0.572 & \textbf{22.459/0.681} \\ \bottomrule
\end{tabular}
\begin{tablenotes}
\item  Pro-UIGAN-C, Pro-UIGAN-M, and Pro-UIGAN-H models are trained on CelebA-HQ, Multi-PIE, and Helen, respectively.
\end{tablenotes} 
\end{threeparttable} 
\label{table-cross}
\end{table}

\subsection{Limitations}
\begin{figure}[t]
\centering
\includegraphics[height=6cm,width=0.45\textwidth]{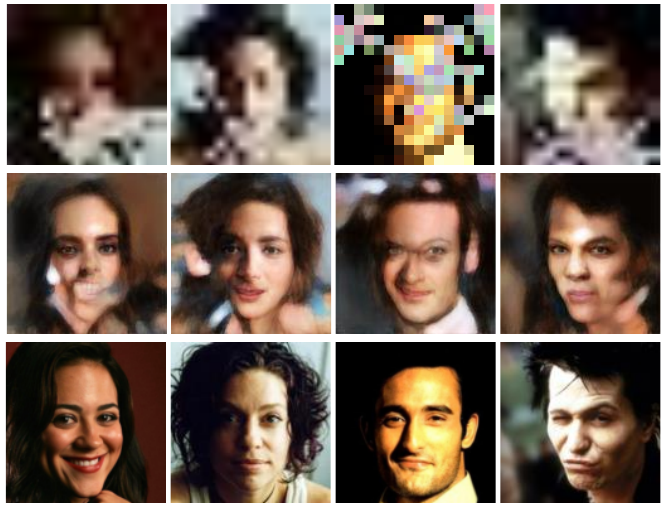}
\caption{Failure cases. First row: input occluded LR faces. Second row: hallucinated results of our model. Third row: Ground-truths.}
\label{limitations}
\end{figure}

Although our Pro-UIGAN model can generate visually appealing non-occluded HR faces, it has limitations.
When hallucinating face images with severe illumination conditions, uneven lighting should yield false landmark detection results. 
This may introduce further ambiguity in the hallucination process.
As shown in Fig.~\ref{limitations}, artifacts and distortions appear on the hallucinated facial appearance.
We will exploit appropriate illumination normalization methods in our future work to make improvements.

\begin{table*}[t]
\caption{Quantitative comparisons on face alignment (NRMSE) and face parsing (IoU).}
\centering
\begin{threeparttable}
\begin{tabular}{@{}c|c|c|c|c|c|c|c|c|c|c|c|cc@{}}
\toprule
\multirow{3}{*}{SR Method} & \multicolumn{2}{c|}{CelebA-HQ} & \multicolumn{2}{c|}{Multi-PIE} & \multicolumn{2}{c|}{Helen} & \multicolumn{2}{c|}{CelebA-HQ} & \multicolumn{2}{c|}{Multi-PIE} & \multicolumn{2}{c}{Helen} \\ \cmidrule(l){2-13} 
                           & \multicolumn{6}{c|}{SR+PA~\cite{li2017generative}}                                                               & \multicolumn{6}{c}{PA~\cite{li2017generative}+SR}                                                               \\ \cmidrule(l){2-13} 
                           & NRMSE           & IoU          & NRMSE           & IoU       & NRMSE           & IoU        & NRMSE           & IoU          & NRMSE           & IoU       & NRMSE           & IoU        \\ \midrule
Bicubic             & 30.67      & 0.0071           & 33.21         & 0.0074       & 35.64        & 0.0076    & 27.31      & 0.0076      & 29.56         & 0.0083           & 32.33        & 0.0082            
     \\ 
FHC~\cite{yu2018face}          & 18.35      & 0.4153        & 20.11         & 0.4049       & 20.87        & 0.3990              & 16.57      & 0.4318         & 17.32         & 0.4229         & 18.26        & 0.4166                 \\
PULSE~\cite{menon2020pulse}        & 40.07      & 0.1740       & 52.08         & 0.0646         & 44.35        & 0.1069              & 38.79      & 0.1834              & 49.22         & 0.0718      & 40.13        & 0.1261   \\ 
Re-CPGAN~\cite{zhang2021recursive}       & 16.24      & 0.4572          & 17.80         & 0.4290          & 19.08        & 0.4105    & 15.31      & 0.4580           & 17.05         & 0.4373          & 18.29        & 0.4152                 \\ \midrule
FCSR-GAN~\cite{cai2019fcsrj}            & 9.74      & 0.5445              & 10.57         & 0.5231       & 11.69        & 0.4870        & 9.74      & 0.5445        & 10.57         & 0.5231       & 11.69        & 0.4870     \\  \midrule
Pro-UIGAN            & 7.16      & 0.6236              & 8.08        & 0.6004    & 9.75      &  0.5803     & 7.16      & 0.6236              & 8.08        & 0.6004    & 9.75      &  0.5803\\
\midrule
Ground-truth HR            & 4.27     &  0.6501             & 2.58        & None      & 3.84       & 0.6056       & 4.27      & 0.6501     & 2.58        & None       & 3.84       & 0.6056\\
\bottomrule
\end{tabular}
\begin{tablenotes}
\item  For Multi-PIE, parsing maps from ground-truth HR images are used as ground-truths.
\end{tablenotes} 
\end{threeparttable} 
\label{table-ap}
\end{table*}

\begin{figure}[t]
\centering
\includegraphics[height=4.3cm,width=0.48\textwidth]{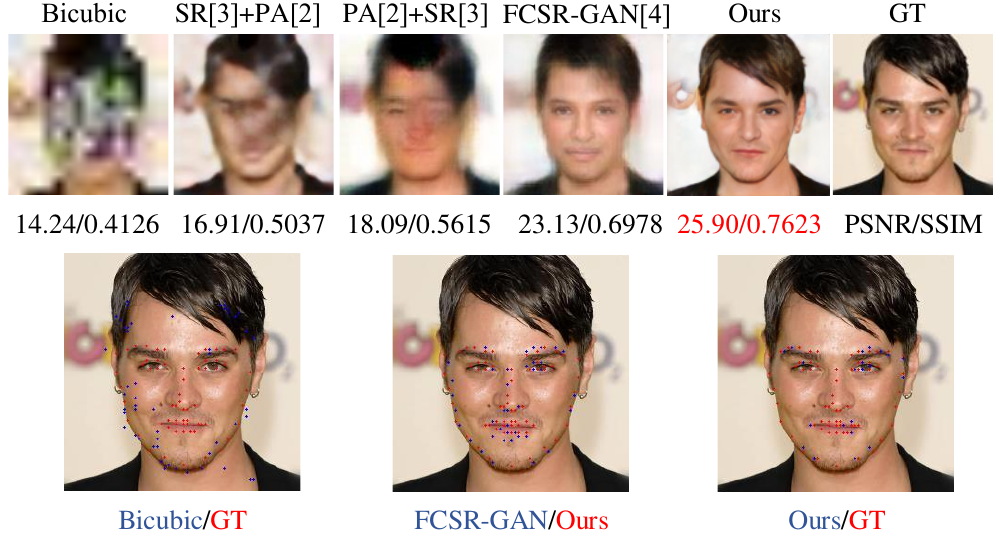}
\caption{Qualitative comparisons of face alignment.}
\label{fig-alignment}
\vspace{-0.4cm}
\end{figure}

\section{Face Hallucination Evaluation via Downstream Tasks}
\subsection{Performance Comparisons on Face Alignment}
We demonstrate that our Pro-UIGAN boosts the performance of low-quality face alignment.
We adopt the ``alignment via hallucination” framework to conduct experiments on the \textbf{CelebA-HQ}, \textbf{Multi-PIE}, and \textbf{Helen} databases.
Specifically, we hallucinate the occluded LR faces first and then use them for face alignment.

Fig.~\ref{fig-alignment} shows the hallucinated images of compared methods and the facial landmarks estimated by FAN~\cite{bulat2017far} on different hallucinated faces.
Moreover, Tab.~\ref{efficiency} provides the NRMSE performance, a commonly used metric in face alignment.
The results indicate that: (1) Our Pro-UIGAN can alleviate the alignment difficulty and thus result in lower NRMSE values (see Tab.~\ref{table-ap}).
(2) In comparison with prior methods, such as FSCR-GAN~\cite{cai2019fcsrj}, our hallucinated face provides visually superior estimation on facial components and shapes.
These results demonstrate that Pro-UIGAN reconstructs facial geometry more accurately and is practical for low-quality face alignment tasks.

\begin{figure}[t]
\centering
\includegraphics[height=12cm,width=0.47\textwidth]{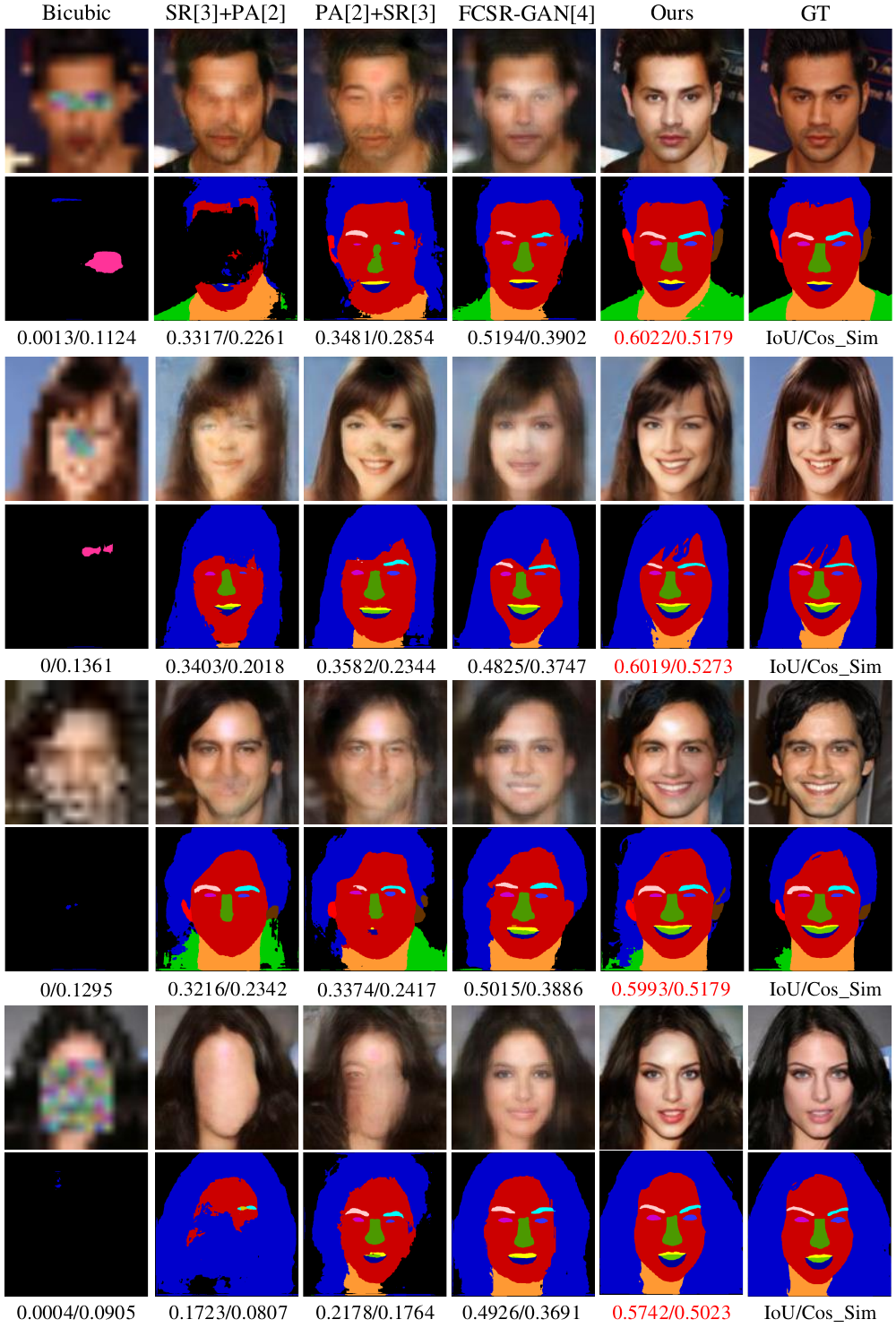}
\caption{Qualitative comparisons of face parsing.}
\label{fig-parsing}
\end{figure}

\subsection{Performance Comparisons on Face Parsing}
We manifest that our Pro-UIGAN also benefits low-quality face parsing tasks.
Similarly, we adopt an off-the-shelf face parsing model~\cite{Liu_2015_CVPR} to conduct face parsing experiments for hallucinated images.

As shown in Fig.~\ref{fig-parsing}, the parsing results of hallucinated face images generated by our method separate complete and accurate facial components.
In contrast, parsing results of hallucinated face images generated by other methods either have wrong shapes or lose key components (\emph{i.e.,} eyes, nose, or mouth).
Meanwhile, we report the Intersection-over-Union (IoU) results to make a quantitative comparison (see Tab.~\ref{table-ap}).
As shown in the table, our Pro-UIGAN yields the highest parsing accuracy and surpasses the competing methods by a large margin (over $5\%$) on all databases.

\subsection{Performance Comparisons on Face Recognition}
In this section, we conduct face recognition experiments to discuss how much our method can benefit the occluded LR face recognition.

\begin{figure}[t]
\centering
\includegraphics[width=0.48\textwidth]{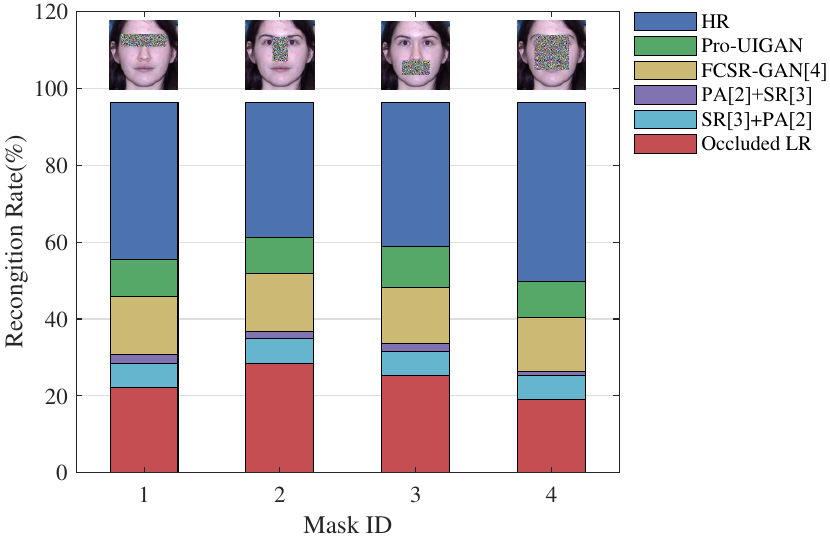}
\caption{Face recognition results on Multi-PIE.}
\label{FR}
\end{figure}

\subsubsection{Experimental settings} 
We use the ``recognition via hallucination” scheme to conduct experiments on \textbf{Multi-PIE}~\cite{gross2010multi}.
First, we split the Multi-PIE face images according to the identity and construct disjoint subject training and testing sets.
Then, we train the competing methods on the training set and conduct face recognition experiments on the testing set.
Here, we choose the frontal HR face images of testing individuals to form our testing set.
We apply the four masking types in Fig.~\ref{fig-parsing} for each HR face.
These masking types simulate the most common occlusions in the real world.
For example, masking mouths simulates wearing masks, and masking eyes mimic wearing glasses.
Finally, we use an off-the-shelf face recognition model~\cite{Liu_2017_CVPR} to conduct face recognition experiments on occluded LR faces and hallucinated non-occluded HR faces generated by compared methods.

\subsubsection{Evaluation}
As shown Fig.~\ref{FR}, the face recognition rate of our hallucinated faces outperforms all other methods, which demonstrates that our Pro-UIGAN possesses better identity preservation ability and benefits low-quality face recognition tasks.

\subsection{Performance Comparisons on Face Expression Classification}
In this experiment, we demonstrate that Pro-UIGAN can preserve facial expressions and boost the performance of low-quality face expression classification.

\subsubsection{Experimental settings}
We conduct a standard 10-fold subject-independent cross-validation~\cite{liu2014facial,liu2014feature} on \textbf{Multi-PIE}~\cite{gross2010multi}.
First, we split the synthesized occluded LR/non-occluded HR Multi-PIE face pairs according to the identity ID and form 10 subject-independent subsets.
Then, we conduct ten-fold validation experiments like ~\cite{liu2014facial,liu2014feature,7961791_FG2017,9471014_tcyb2021}.
We adopt an off-the-shelf expression classification model~\cite{albanie16learning} to classify the facial expression of hallucinated faces, report the average of classification scores for the 10 subsets.
Here, we set the classification results of the occluded LR faces and the non-occluded HR ones as the lower and upper bounds in this task (see Tab.~\ref{tablefsr}).

\subsubsection{Evaluation}
\begin{figure}[t]
\centering
\includegraphics[width=0.48\textwidth]{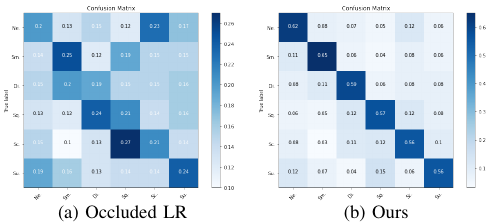}
\caption{The confusion matrices of LR faces and our hallucinated ones. Darker colors indicate higher accuracy.}
\label{figfer}
\end{figure}

As shown in Tab.~\ref{tablefsr}, the non-occluded HR faces hallucinated by our Pro-UIGAN achieve superior expression classification accuracy compared to the other methods.
In the confusion matrix for occluded LR faces and our hallucinated ones (see Fig.~\ref{figfer}), 
it can be observed that the faces hallucinated by Pro-UIGAN outperform occluded-LR ones on all the expressions.
These results indicate that Pro-UIGAN recovers authentic facial expressions and benefits low-quality face expression classification tasks.

\begin{table}[t]
\caption{Face expression classification results on Multi-PIE.}
\centering
\begin{tabular}{@{}c|c|cc@{}}
\toprule
\multirow{2}{*}{SR method} & \multicolumn{2}{c}{Accuracy} \\ \cmidrule(l){2-3} 
                           & PA+SR          & SR+PA          \\ \midrule
Bicubic                    & 22.34\%       & 20.59\%        \\ 
PULSE~\cite{menon2020pulse}                        & 27.40\%       & 24.16\%             \\ 
FHC~\cite{yu2018face}                       & 30.66\%       & 28.31\%   \\ 
Re-CPGAN~\cite{zhang2021recursive}                      & 35.07\%       & 33.92\%          \\ \midrule
FCSR-GAN~\cite{cai2019fcsrj}             & \multicolumn{2}{c}{45.93\%}\\ \midrule
Occluded LR             & \multicolumn{2}{c}{22.14\%}  \\ \midrule
Non-occluded HR                 & \multicolumn{2}{c}{94.62\%}  \\ \midrule
Pro-UIGAN                  & \multicolumn{2}{c}{\textbf{56.31\%}}  \\ \bottomrule
\end{tabular}
\label{tablefsr}
\end{table}

\section{Conclusion}
This paper presents a Pro-UIGAN framework to jointly super-resolve and inpaint occluded LR face images.
Our developed multi-stage progressive learning strategy allows us to achieve coarse-to-fine feature refinement under the guidance of facial geometry priors, significantly improving the hallucination performance.
Experimental results on multiple widely-used benchmarks confirm the efficacy of Pro-UIGAN in generating identity-preserving face images and substantially benefiting downstream tasks.

\section*{Acknowledgment}
The authors would like to thank the anonymous reviewers and the Associate Editor for their helpful suggestions and valuable comments.


\bibliographystyle{IEEEtran}

\end{document}